\documentclass[pdflatex, sn-mathphys]{sn-jnl}
\usepackage[utf8]{inputenc}
\usepackage[T1]{fontenc}
\usepackage{graphicx}
\usepackage{amsmath}
\usepackage{booktabs,multirow}
\usepackage{array}
\usepackage{xcolor}
\usepackage{float}
\usepackage{fancyvrb}
\usepackage{cleveref}
\usepackage[numbers,sort&compress]{natbib}
\usepackage{mdframed}
\usepackage{fvextra}
\usepackage{array}
\newcolumntype{L}[1]{>{\raggedright\arraybackslash}m{#1}}
\newcolumntype{C}[1]{>{\centering\arraybackslash}m{#1}}

\begin{document}

\title{LLM-as-a-Judge is Bad, Based on AI Attempting the Exam Qualifying for the Member of the Polish National Board of Appeal}
\author{Michał Karp}
\author{Anna Kubaszewska}
\author{Magdalena Król}
\author{Robert Król}
\author{Aleksander Smywiński-Pohl}
\author{Mateusz Szymański}
\author{Witold Wydmański}
\date{October 2025}

\abstract{
  This study provides an empirical assessment of whether current large language models (LLMs) can pass the official qualifying examination for membership in Poland’s National Appeal Chamber (Krajowa Izba Odwoławcza). The authors examine two related ideas: using LLM as actual exam candidates and applying the 'LLM-as-a-judge' approach, in which model-generated answers are automatically evaluated by other models. The paper describes the structure of the exam, which includes a multiple-choice knowledge test on public procurement law and a written judgment, and presents the hybrid information recovery and extraction pipeline built to support the models. Several LLMs (including GPT-4.1, Claude 4 Sonnet and Bielik-11B-v2.6) were tested in closed-book and various Retrieval-Augmented Generation settings. The results show that although the models achieved satisfactory scores in the knowledge test, none met the passing threshold in the practical written part, and the evaluations of the 'LLM-as-a-judge' often diverged from the judgments of the official examining committee. The authors highlight key limitations: susceptibility to hallucinations, incorrect citation of legal provisions, weaknesses in logical argumentation, and the need for close collaboration between legal experts and technical teams. The findings indicate that, despite rapid technological progress, current LLMs cannot yet replace human judges or independent examiners in Polish public procurement adjudication. }
  
\keywords{Large Language Models (LLMs), legal reasoning, public procurement law, National Appeal Chamber (Poland), legal exam evaluation, Retrieval-Augmented Generation (RAG), LLM-as-a-judge}

\maketitle

\section{Introduction}


With the growing popularity of Large Language Models (LLMs) it is observed that people want these 
Artificial Intelligence (AI) models to accomplish more sophisticated tasks. One of the areas that 
is particularly well suited for taking advantage of them is law. This domain of human endeavor is
dominated by text and thus is particularly well suited for receiving help from LLMs. Inspired by
this observation, in this research we want to find out if an LLM can pass an exam for a judge in the
Polish National Appeal Chamber (pol. \textit{Krajowa Izba Odwoławcza, NAC}), verifying its competence
in a well defined testing procedure where theoretical knowledge of provisions has to be combined
with practical ability to apply these laws in a legal case.

On the other hand, the term LLM-as-a-judge \cite{Gu2025LLMasJudgeSurvey}, functions as a name for the procedure when an LLM is applied
to judge if a text generated by this or another LLM is valid vis-a-vis certain reference answer. 
The idea to follow such a procedure stems from two facts: it is impossible to apply well-defined 
metrics such as precision and recall to a generated content and that human evaluation of generated content
is time consuming and costly. One solution is to fed the LLM-as-a-judge with the generated content and the
reference content and ask it, how much the generated content reflects the features of the reference answer.

In this research we show that combining these two approaches in the context of law may lead to unexpected an
unwanted outcomes. In the rest of the paper we introduce the exam for judges in the Polish National Appeal Chamber,
how this exam was approached with the use of LLMs and what were the outcomes of the preliminary experiments
we have conducted to evaluate the probability of the models passing the exam. Then we describe the actual exam
taken by the LLMs in the same regime as the human participants and compare the assessment made
by the real examining committee with the assessment made by LLM-as-a-judge system. We show that
although the metrics achieved by our models in the preliminary experiments were pretty satisfactory,
the final exam was not passed by any of the models.

\subsection{Research Motivation and Organizational Impact}

The Public Procurement Office, which provides operational support to the NAC, initiated this research to evaluate the extent to which Large Language Model (LLM) technology could support the work of Chamber adjudicators in various capacities including charge synthesis, documentation searches, identification of precedential rulings in analogous factual circumstances, legal provision analysis for identifying adjudicatory trends, and general knowledge assessment within the specialized field of public procurement law. The experimental results will inform management decisions regarding the scope of further implementation of this technology within the institutional framework.

The potential positive impact of successful examination performance by AI systems on the organization extends beyond technological validation. Such outcomes could demonstrate the feasibility of AI-assisted legal research tools, potentially enhancing the efficiency and consistency of adjudicatory processes while maintaining the high standards of legal analysis required in public procurement dispute resolution. This technological advancement could contribute to more streamlined case processing and improved accessibility of legal reasoning patterns, ultimately serving the broader goal of ensuring fair and efficient public procurement procedures in Poland.

\section{Exam for a Judge in the National Appeal Chamber}

\subsection{National Appeal Chamber}
The National Appeal Chamber is a specialized adjudicatory body established within the Polish public procurement system, operating as the primary instance for resolving procurement-related disputes.
The Chamber serves as the competent authority for hearing appeals submitted in public procurement proceedings and replaced the previous system of arbitration panels that had operated since 2007.

\subsection{Scope of Competence and Legal Framework}
The Chamber's primary responsibilities encompass: (1) hearing appeals in cases referenced in Article 513 of the Public Procurement Law (\textit{Prawo Zamówień Publicznych}) \cite{PPL2019}; (2) reviewing applications for lifting contract conclusion prohibitions as specified in Article 578(1) of the Public Procurement Law; and (3) adopting resolutions containing opinions on contracting authorities' objections to the results of preliminary and ad-hoc controls conducted by the President of the Office.
The legal foundation for the Chamber's operations is established in Articles 473-492 of the Public Procurement Law. The Chamber comprises no more than 100 members, appointed by the minister responsible for economic affairs from among candidates meeting statutory requirements who achieved the best results in qualification proceedings. Members maintain independence in their adjudication and are bound exclusively by applicable legal provisions.

\subsection{Qualification Process Structure for Chamber membership}

The recruitment process for Chamber membership follows a rigorous three-stage assessment protocol regulated by the Regulation of the Prime Minister of February 22, 2021, regarding the procedure for conducting qualification and supplementary qualification proceedings for candidates for National Appeal Chamber membership \cite{RMF2020}. The examination structure comprises: (1) a knowledge assessment consisting of 50 single-choice questions testing theoretical understanding of public procurement and civil law, worth a maximum of 100 points; (2) a written practical examination requiring legal analysis of a factual case study within a 120-minute timeframe, also worth 100 points; and (3) a qualification interview evaluating personal characteristics and professional competencies, worth up to 30 points.

\subsection{The 2025 Recruitment Cycle}
The most recent qualification proceeding was announced on March 7, 2025, by the Minister of Development and Technology, acting pursuant to Article 477(6) of the Public Procurement Law and Section 18(1) of the Prime Minister's Regulation of February 22, 2021. The recruitment cycle established a limit of 13 positions to be filled through the qualification process, with application submissions accepted until March 3, 2025.

The qualification requirements mandate candidates to possess: Polish citizenship; completion of legal studies; successful completion of judicial, prosecutorial, notarial, attorney, or legal counsel examinations; minimum five years of professional experience in public procurement and civil law; full legal capacity; enjoyment of full public rights; clean criminal record; and appropriate security clearance for accessing classified information.

\subsection{Required Legal Knowledge Areas}

The qualification examination encompasses a comprehensive range of legal domains essential for effective adjudication in public procurement matters. The statutory scope of required knowledge, as defined in Annex 1 to the Prime Minister's Regulation, includes fourteen distinct areas of legal expertise.

The primary focus centers on Polish public procurement legislation, which includes the Public Procurement Law, the Concession Act \cite{CL2016}, their implementing regulations, and the development of jurisprudence through decisions of the Supreme Court, district courts, Constitutional Tribunal, and the Chamber itself. Candidates must demonstrate comprehensive understanding of European Union law governing public procurement and concessions, including fundamental Treaty principles, key directives (2014/24/EU, 2014/25/EU, 2014/23/EU, 2009/81/WE), remedies directives, the Common Procurement Vocabulary regulation, and the jurisprudence of the Court of Justice of the European Union.

Essential civil law competencies include mastery of the Civil Code, Civil Procedure Code (excluding enforcement and interim measures), Construction Law, and criminal law provisions concerning economic crimes. The examination also requires knowledge of public finance principles, including organizational forms of public finance sector entities, financial management rules, and liability for violations of public finance discipline.

Administrative law components encompass the Code of Administrative Procedure, organizational principles of governmental administration and local self-government, while commercial law knowledge includes the Commercial Companies Code and unfair competition regulations. Specialized areas include public-private partnership law, intellectual property law covering copyright scope and protection, and classified information protection regulations.

\section{Related Work}

From the perspective of our experiments, the most relevant point of reference comes from prior work on professional bar examinations in the United States and Brazil. These exams are not only high-stakes gateways to legal practice, but also carefully designed assessments of both doctrinal knowledge and applied reasoning, which makes them particularly instructive for evaluating the capabilities of large language models \cite{Katz2024GPT4BarExam}\cite{Martinez2024ReEvaluatingGPT4UBE}\cite{FreitasGomes2023OAB}.

In the U.S., research has centered on the Uniform Bar Examination (UBE), which is divided into three parts: the multiple-choice Multistate Bar Examination (MBE), the essay-based Multistate Essay Examination (MEE), and the Multistate Performance Test (MPT), which simulates real-world legal tasks such as drafting memoranda or client letters. Early studies reported that GPT-4 outperformed the average human examinee on the MBE, suggesting that the model had mastered a wide range of doctrinal material \cite{Katz2024GPT4BarExam}. Performance on the essay and practical sections, however, was more difficult to evaluate, as these tasks required structured argumentation and professional writing. Human examiners had to be brought in to ensure fair and consistent grading. Later re-evaluations cautioned that whether GPT-4 could truly be said to “pass” the bar depended heavily on the scoring rubric and scaling methodology used. When more conservative mappings to the official score distribution were applied, the model’s performance fell closer to the 40–60th percentile of human test-takers, tempering initial claims of equivalence with licensed practitioners \cite{Martinez2024ReEvaluatingGPT4UBE}.

A comparable experiment was conducted in Brazil with the Exame da Ordem dos Advogados do Brasil (OAB), which similarly unfolds in two stages. The first is an 80-question multiple-choice test covering broad areas of law, while the second requires the drafting of a procedural petition and answers to open-ended problem questions within a chosen legal domain. Here, too, LLMs performed well in the objective phase, surpassing the threshold required to advance \cite{FreitasGomes2023OAB}. But in the second stage, which demands argumentation, statutory interpretation, and adherence to formal requirements of Brazilian legal practice, the models underperformed significantly. Legal specialists judged their submissions to be incomplete, incoherent, or procedurally deficient, confirming that the gap between factual recall and professional reasoning remains substantial.

These findings from the U.S. and Brazil illustrate a broader pattern: LLMs demonstrate surprising competence on knowledge-based, multiple-choice questions but struggle with open-ended tasks that require structured reasoning, practical judgment, and procedural accuracy. They also underscore the indispensable role of human evaluators, both in grading free-response answers and in calibrating what it means to “pass” under official standards.

A growing body of research has generalized these lessons to other professional and educational benchmarks. Early studies on the UBE showed that GPT-4 could exceed human averages on doctrinal knowledge, while later critiques emphasized the fragility of such claims under different scoring assumptions \cite{Katz2024GPT4BarExam, Martinez2024ReEvaluatingGPT4UBE}. Similar observations were made for the Brazilian OAB exam \cite{FreitasGomes2023OAB} and for other high-stakes tests such as the LSAT, where models falter on logic-intensive reasoning tasks \cite{Malik2024LSATLogicGames}.

These results align with methodological insights from broader evaluation frameworks. Benchmarks like MMLU \cite{Hendrycks2021MMLU}, AGIEval \cite{Zhong2023AGIEval}, and evaluations of GPT-4 on the USMLE \cite{Nori2023GPT4Medical} highlight the importance of transparent reporting—covering inference settings, explicit metrics, and percentile mappings—as well as careful calibration of uncertainty. Without such standards, claims that LLMs “pass” exams are easily overstated or irreproducible.

Another recurring lesson is the necessity of human oversight. 
Beyond domain-specific exams, automatic LLM-as-judge evaluation has been explored with G-Eval, which uses chain-of-thought and a structured form-filling rubric to rate model outputs; it reports substantially higher correlations with human raters, while also noting biases of LLM judges toward LLM-generated text \cite{Liu2023GEval}. In the legal domain, a preregistered empirical study of proprietary RAG-based legal research tools (Lexis+ AI, Westlaw AI-Assisted Research, Ask Practical Law AI) finds hallucination rates of 17–33\% and wide variation in responsiveness and accuracy—undercutting “hallucination-free” claims and reinforcing the need for human verification and careful evaluation design \cite{Magesh2024HallucinationFree}.
Tasks involving legal reasoning, statutory interpretation, or judgment writing cannot be reliably graded by LLMs alone. This is particularly evident in the growing literature on “LLM-as-a-Judge” approaches, which explore the use of models as automated evaluators of other models’ outputs. While attractive for their scalability, these methods have been shown to introduce systematic biases, such as favoring verbose answers, over-weighting the order in which options are presented, and producing inconsistent scores depending on prompt framing. Surveys of the field emphasize that these biases cannot be eliminated simply by scaling models, and instead require corrective mechanisms such as expert annotation, majority-vote consensus across multiple models, or hybrid pipelines in which human judgment plays a decisive role \cite{Gu2025LLMasJudgeSurvey}. The experience from bar exam evaluations confirms these concerns: where grading depends on argumentative quality and legal reasoning, human examiners remain indispensable.

Closely related is the problem of data contamination: performance may be artificially inflated if exam items have leaked into training corpora. Recent work has documented this issue across multiple benchmarks and calls for systematic validation of experimental results \cite{Dekoninck2024ConStat}.

Finally, specialized legal benchmarks have been developed to go beyond general-purpose knowledge testing. Datasets such as LegalBench \cite{Guha2023LegalBench}, LawBench \cite{Fei2023LawBench}, and the Reasoning-Focused Legal Retrieval benchmark \cite{Zheng2025RFLegalRetrieval} offer tasks curated by legal experts that emphasize statutory reasoning and applied legal analysis. In parallel, holistic evaluation frameworks such as HELM \cite{Liang2023HELM} advocate multidimensional assessment that accounts not only for accuracy, but also calibration, robustness, fairness, and efficiency. Together, these strands of research suggest that future evaluations of LLMs on bar exams—and legal reasoning more generally—must integrate retrieval, reasoning, calibration, and systematic human-in-the-loop validation to yield a reliable picture of model capabilities.

\section{Research Questions}
The primary motivating factor of our research was to find out to what extent LLMs can 
pass the exam. But such a crude question can be broken down into smaller, more
detailed questions presented below:

\begin{enumerate}
    \item[1.] Can AI pass the exam qualifying for the member of the Polish National Appeal Chamber?
    \item[2.] How good are AI models in answering direct legal questions related to the Polish public procurement?
    \item[3.] Which models are the best in assessing formal requirements of the legal procedure and are they good enough to perform such an assessment?
    \item[4.] Which models stand out in summarizing the legal research in a form of a judgment and can they be employed as a tool facilitating judgment writing?
    \item[5.] Can an LLM-as-a-judge be employed to assess the validity of the exam?
\end{enumerate}

Regarding the first question, that was our primary motivating force. We wanted to know 
if it is already feasible for the Polish language and Polish jurisdiction to apply
LLMs currently, like it was done for English (for USA legislation) and Brazilian-Portuguese
(for Brazilian legislation). 

The second question relates to the fact that the exam consists of two parts. The first one
is concerned with the knowledge of the law related to the public procurement. It is 
a test with just one valid answer out of 3, so a we could test a typical Retrieval Augmented
Generation (RAG) scenario and also closed-book scenario, when the only source of knowledge
is the model itself.

The third question relates to the fact that when the cases is decided in KIO (or any other court), there is a large
number of formal requirements that have to be validated that influence various aspects
of the decision, such as the type of decision or if the factual matters should ever be 
evaluated, e.g. if the protest was submitted too late. Since these factors take the form
of binary or fixed-set-outcome decisions, they can be evaluated with well defined, machine 
learning measures such as precision and recall. By extracting them from the general
decision-making process, we would be able to see if there are areas where the models
can be safely applied, with limited risk of spectacular failure.

The fourth research question was directly related to the first one, but it concentrates
on the assessment of the ability of the LLMs to write a judgment. This seems to be the 
most interesting part of the whole experiment, since if we find that one of the models 
excels in this task, we can hope that at least one model will pass the exam and in a 
broader context, can be used to facilitate the  judgment process for KIO.

The last question is concerned with the major approach to testing LLM generated
content. Currently, methods such as LLM-as-a-judge \cite{Gu2025LLMasJudgeSurvey} are 
applied in a growing body of research. They are based on the assumption that when 
the judge LLM is given the reference answer and the answer generated by judged LLM,
the judge can assess -- on par with a human evaluator -- the correctness of the
generated answer. This approach stems from the fact that manual evaluation is 
expensive, especially when the amount of output is huge. However in the case
of law, we are not aiming at ``correct on average'', but ``always correct''
paradigm. So we wanted to see how far away the LLM used as a judge is from the
assessment given by human evaluators in an exam designed for people.

\section{Materials and Methods}

\subsection{Data from Previous Exams}

To gain a deeper understanding of the task and to adapt our methods to the 
specific requirements of the exam, we conducted preliminary experiments using
publicly available data from previous exam editions. In total, we collected
data from five past exams, comprising 250 questions from the knowledge tests 
and five case studies for the judgment-writing section. Using these materials, 
we designed and optimized processing pipelines that achieved the highest performance 
on this benchmark data. During the actual exam (cf. \ref{subsec:exam}), only these
optimized pipelines were employed, together with the selected language models (see \ref{sec:llms} for details).

\subsection{Large Language Models}
\label{sec:llms}

The study evaluated three large language models (LLMs), differing in scale, training approach, and intended use:
\begin{itemize}
\item \textbf{Bielik-11B-v2.6-Instruct} (\citet{ociepa2024bielik7bv01polish}) is the latest Polish-language open-source model released by the SpeakLeash team, containing 11 billion parameters. It was trained primarily on Polish data and fine-tuned (instruct-tuning) for conversational and instruction-following tasks. In our experiment it served as a native Polish model, allowing us to assess how a locally developed and deployed LLM compares with global state-of-the-art models in legal tasks. 

\item \textbf{Claude 4 Sonnet}\footnote{\url https://www.anthropic.com/news/claude-4} is a part of Anthropic’s Claude family, designed as a balance between speed and output quality. This is a general-purpose, multilingual model that performs well in complex analytical and legal-reasoning tasks, though it is not specifically adapted to Polish law.

\item \textbf{GPT-4.1}\footnote{\url{https://openai.com/index/gpt-4-1/}} was the latest iteration in OpenAI’s GPT at the time of the study, in the same time it was considered the most advanced general-purpose model available. It offers strong capabilities for long-context processing, complex reasoning, and high-quality text generation across multiple languages, including Polish.
\end{itemize}

All three models were subjected to the same testing procedures described in the article: a statutory-knowledge multiple-choice test and a practical written exam requiring the drafting of a legal judgment. This setup enabled a direct comparison between the domestically developed Bielik model and leading commercial LLMs (Claude 4 Sonnet and GPT-4.1) in the context of the Polish National Appeal Chamber qualification exam.

\subsection{Advanced RAG Pipeline}
\label{sec:advanced}

The foundation of our research system rests upon a comprehensive data acquisition and preprocessing pipeline (\textit{advanced RAG pipeline}) that systematically transforms raw legal texts from multiple authoritative sources into a vectorized, searchable repository. This methodology encompasses both the collection of diverse legal documents and their subsequent processing to ensure consistent representation while preserving semantic richness and legal context essential for effective information retrieval.

\subsubsection{Data Collection Strategy}

Our system incorporates legal documents from three distinct and complementary sources, each requiring specialized collection methodologies due to varying data formats and access mechanisms. The data acquisition process was designed to ensure comprehensive coverage of Polish public procurement law from multiple authoritative perspectives.

The \textbf{sentences} collection derives from raw judicial decisions and court judgments obtained through systematic downloads from official FTP servers in PDF format. These documents represent the practical application of legal principles in real-world scenarios and offer insights into judicial decision-making processes. The PDF files underwent optical character recognition and text extraction processes via Docling \cite{Docling} software to convert them into machine-readable formats suitable for further processing.

The \textbf{summaries} collection comprises publicly available analytical summaries and commentaries sourced from the official government portal\footnote{\url{https://www.gov.pl/web/uzp/orzecznictwo-dotyczace-zamowien-publicznych}}. These documents were downloaded in DOCX format and required specialized parsing to extract the textual content while preserving the document structure and metadata. The summaries provide structured overviews of significant legal developments and regulatory interpretations in a more condensed format than full judicial decisions.

The \textbf{PZP} collection encompasses comprehensive official commentary on the Public Procurement Law (\textit{Prawo Zamówień Publicznych}) obtained through automated web scraping of the specialized legal commentary platform\footnote{\url{https://ekomentarzpzp.uzp.gov.pl/prawo-zamowien-publicznych}}. This collection required systematic retrieval of commentary for each article of the Public Procurement Law (articles 1 through 623) through automated API interactions with the commentary platform, followed by HTML parsing and text extraction to eliminate markup artifacts while preserving substantive commentary content.

\subsubsection{Document Processing Pipeline}

Following data collection, all documents underwent a standardized preprocessing pipeline designed to ensure consistency across different source formats and document types. The processing methodology transforms the collected raw documents into a uniform representation suitable for vectorization and search operations.

\paragraph{Text Segmentation and Chunking Strategy}
The preprocessing pipeline employs a straightforward token-based segmentation approach designed to balance computational efficiency with embedding model constraints. Raw documents underwent systematic chunking using the SentenceTransformers-based token text splitter, configured with a fixed maximum chunk size of 256 tokens to ensure compatibility with the underlying embedding models' input limitations. This chunking process utilizes the \texttt{sdadas/mmlw\--retrieval\--roberta\--large} \cite{dadas2024pirb} model's tokenization scheme to ensure consistency between the preprocessing and vectorization stages, creating uniformly sized text segments that can be efficiently processed by the vector embedding pipeline.

\paragraph{Temporal Information Extraction and Document Identification}
Each processed document underwent temporal analysis to extract publication dates using pattern matching techniques specifically designed for Polish date formats. The system employed regular expressions to identify and parse dates in the canonical Polish legal format, converting these temporal markers into standardized epoch timestamps for efficient chronological indexing and retrieval. Document uniqueness was ensured through cryptographic hashing using SHA-256 algorithms applied to the full document text, creating unique identifiers that prevent duplication while enabling efficient existence checking during the upload process.

\paragraph{Quality Control and Filtering Mechanisms}
The preprocessing pipeline incorporated quality control measures to ensure that only substantive content enters the knowledge base. Document chunks underwent length validation, with segments containing fewer than 100 characters being excluded from the final collection to eliminate low-information content such as headers, footers, or fragmented text segments. Additionally, the system implemented deduplication mechanisms at both the document and chunk levels, preventing redundant information from artificially inflating search results or degrading retrieval performance.

This comprehensive data collection and preprocessing methodology ensured that our document repository maintains high standards of content quality, temporal accuracy, and semantic richness while providing the vectorized foundation necessary for effective hybrid search operations across diverse legal document types.

\subsection{Hybrid Search Architecture and Information Retrieval}
\label{sub:search}
The information retrieval component employs a hybrid search architecture that integrates semantic vector-based search with traditional keyword-based text matching. This approach ensures comprehensive coverage of user information needs while maintaining high precision across diverse query types and legal contexts.

\subsubsection{Query Processing and Vectorization Strategy}

The search process begins with domain-specific query preprocessing designed to optimize semantic understanding and retrieval effectiveness. Incoming queries undergo contextual enhancement through prefix application that signals search intent to the vectorization model. Queries are augmented with the prefix \texttt{"[query]: "} to align with the model's training paradigm and enhance semantic comprehension in legal contexts.

Enhanced queries are transformed into high-dimensional vector representations using the specialized \texttt{sdadas/mmlw-retrieval-roberta-large-v2} \cite{dadas2024pirb} embedding model designed for multilingual legal document retrieval without pooling mechanisms. This model has been optimized for cross-lingual legal text comprehension and provides robust semantic understanding across various linguistic and semantic patterns within legal queries.

\subsubsection{Multi-Collection Search Architecture}

The search system operates across a multi-collection architecture organizing legal documents by content granularity and document type. The \textbf{sentences} collection contains fine-grained segments from judicial decisions enabling precise retrieval of specific legal reasoning. The \textbf{summaries} collection provides higher-level analytical content facilitating broader contextual information retrieval. The \textbf{PZP} collection offers comprehensive statutory commentary supporting detailed regulatory research.

Each collection maintains parallel storage in Typesense \cite{noauthor_typesensetypesense_2025} with both full-text indexing for keyword matching and vector representations for semantic similarity search. Corresponding full-document collections preserve complete textual content, enabling comprehensive context retrieval when detailed analysis is required.

\subsubsection{Hybrid Search Execution and Ranking}

The core search execution employs a hybrid methodology that simultaneously leverages multiple retrieval signals to optimize result relevance. The system performs parallel execution of semantic vector similarity matching and traditional keyword-based text search across relevant document collections.

Semantic search utilizes cosine distance calculations between query vectors and document embeddings to identify conceptually related content that may lack exact terminological overlap. Keyword-based search ensures exact term matching for precise legal terminology and specific regulatory references requiring literal correspondence. Typesense's integrated hybrid reranking mechanism combines both signals, applying scoring algorithms that weight semantic similarity and textual relevance according to query characteristics.

The search process initially retrieves an expanded result set (five times the requested number) to accommodate subsequent deduplication and ranking optimization. Retrieved results undergo post-processing including intelligent deduplication that consolidates multiple segments from identical source documents by retaining only the segment with lowest vector distance. Final ranking considers vector similarity scores, keyword matching relevance, document metadata, and collection source information.

\subsubsection{Mixed Collection Search and Cross-Domain Retrieval}

For comprehensive coverage, the system supports mixed collection search enabling simultaneous querying across all document types. Mixed searches execute parallel queries across \textit{sentences}, \textit{summaries}, and \textit{PZP} collections while maintaining collection provenance metadata. Cross-collection deduplication prevents redundant information retrieval while preserving perspective diversity. Final result aggregation combines and ranks results from all collections using unified scoring criteria, ensuring relevant information surfaces regardless of source collection while maintaining balanced representation across different document types.

This hybrid search architecture effectively addresses diverse legal information needs while maintaining high standards of relevance, precision, and coverage across Polish public procurement law.

\subsection{RQ2 -- Knowledge of Regulations}

In order to evaluate the feasibility of the models to pass the first part of the exam,
which is concerned with the knowledge of the Polish law,
we have investigated three approaches:
\begin{enumerate}
    \item Closed-book -- an LLM is given the question and the possible answers and selects
    one of the answers based on it's internal knowledge, without any supporting material;
    \item Simple RAG -- the content of bills and ordinances is loaded to a vector store
    and searched for each question, the model is given that information as the context;
    \item Advanced RAG -- this approach uses the same pipeline as for the legal research
    (explained in \Cref{sub:search}) and is given to the model as the context when determining the 
    correct answer.
\end{enumerate}

The first approach is our baseline, regarding the knowledge test. It is 
straightforward, so we only give the details regarding the prompt,
the model parameters and the way the answer is extracted from the generated text. We have 
used a basic 
regular expression to extract the letter indicating the answer from the output generated
by the model: \verb+r'\b([abc])\b'+. This code matches the first occurrence of \textit{a},
\textit{b} or \textit{c} in the generated text. 
The prompt used to generate the answer is given in Listing \ref{listing:knowledge-test-prompt}.
For GPT and Claude we have used default parameters with temperature set to 0 and number of completion
tokens set to 10. For Bielik we have used temperature 0.01 and number of completion 
tokens set to 1000. The reason for these parameters change was two-fold. First of all, the
configuration of the model with temperature set to 0, produced null response, thus it was raised 
to a very small, positive number, which fixed that problem. Regarding the number of tokens --
despite the direct instruction, the model almost always produced some justification for the answer.
So in order to give it some space to reason about the answer and not to penalize it for that
reasoning, we gave it much more tokens.

\begin{figure}[t]
\centering
\begin{Verbatim}
Pytanie: {question}                  Question: {question}

Opcje odpowiedzi:                    Answer options:
a) {option_a}                        a) {option_a}
b) {option_b}                        b) {option_b}
c) {option_c}                        c) {option_c}

Odpowiedz tylko jedną literą         Answer with one letter only
(a, b lub c) wskazującą              (a, b or c) indicating 
prawidłową odpowiedź.                the correct answer.
\end{Verbatim}
\caption{LLM Prompt for Question Evaluation -- original (Polish) and translated to English. The Polish
version was used for all LLMs.}
\label{listing:knowledge-test-prompt}
\end{figure}

The second approach was more involved. The text of the relevant bills was divided by 
\textit{articles}  (the primary organizing element for bills) or \textit{paragraphs} 
(the primary organizing element for ordinances) and loaded to a vector store.
Regarding the indexing of provisions, once they were divided by articles or paragraphs,
they were vectorized using the Qwen 3, 0.6B embedding \footnote{\url{https://huggingface.co/Qwen/Qwen3-Embedding-0.6B}} model.
If the full content of the article
did not fit the length limit of that model (8192 tokens), the vector representation was
created on the truncated version of the provision. Since the number of provision
was not very large, we have compared the question vector with all provision vectors,
to determine those, most relevant for the question. We have tried two approaches -- one where only
the ,,question-like'' part of the question was used as the query, and the other when the possible
answers were also included. The model was instructed with the following query (for the version with 
possible answers):
\texttt{Given the beginning of the text (question-like) and the possible answers, retrieve 
the most relevant documents that will help answer the question.}
Such a context including \textit{k=10}
provisions was provided with the question and the possible answers to the model,
in order to decide which answer is the most probable.

The last approach was the most involved. It used the database with provisions and 
judgments indexed in the Typesene vector store (as described in Section \ref{sub:search}). 
Here we took only the question-like part (without the possible answers) of the
question-answers structure to reformulate it into a real question used in the research pipeline. 
Once the processing was finished, the relevant context together with the question and the 
possible answers were given to the model to decide which of the answers is
the most probable.

\subsection{RQ3 -- Formal Assessment}
\label{sec:formal}

The written part of the exam consists of writing a judgment or decision for a given set of facts. In the written work, we distinguished between a formal assessment (whether the formal criteria were met by the appellant) and a substantive assessment (whether the appellant's allegations were justified). 

The first step the judge takes before proceeding to the merits of the case is to verify the compliance of the appeal with the formal criteria and to decide whether the case should be adjudicated by a ruling or a judgment. According to the Public Procurement Law, an appeal may be rejected due to failure to meet formal criteria, may be withdrawn by the appellant, or the appeal proceedings may be discontinued. In this section, we present a simple and complete architecture that enabled the assessment of the case description in terms of the criteria described.

\begin{figure}[b]
\centering
\begin{Verbatim}
## 1. Entry in proceedings for supplies, services, or a competition: 
entry **{{ publishedInBZP? "7500 PLN" : "15000 PLN" }}**
---
## 2. Entry in proceedings for construction works: 
entry **{{ publishedInBZP? "10000 PLN" : "20000 PLN" }}**
---
## 3. Entry in proceedings for social services and other special 
services: entry **{{ publishedInBZP? "7500 PLN" : "15000 PLN" }}**
---
## 4. When the case concerns the fact that the Contracting Authority 
should have organized a proceeding or competition, but did not.
* **Supplies, services, or competition** → entry **7500 PLN**
* **Construction works** → entry **10000 PLN**
---
## 5. Single entry for partial offers

If you appeal against an act or omission concerning several
parts of the contract, you submit **one entry**.
\end{Verbatim}
\caption{A simplified version of § 2 of the Regulation of the Minister of Finance used in determining the expected fee. The text was originally in Polish, but it has been translated for the sake of linguistic consistency in this article.}
\label{listing:simplified-regulation}
\end{figure}

\subsubsection{Information extraction pipeline}
In the case of a judicial examination, a precise description of the facts is provided. Relevant details are clearly indicated, while irrelevant information is often omitted. The information described below is extracted from such a description. However, in real cases, the source of information is a collection of legal documents from which the information must be taken directly.

\begin{figure}[t]
\centering
\begin{Verbatim}
{
  "type": "object",
  "properties": {
    "numOfUnits": {
      "type": "number",
      "description": "The number of days or months that the Appellant 
      has to file an appeal from the date of the action."
    },

    "unitName": {
      "type": "string",
      "description": "Only one of two options - \"day\" or \"month\""
    },

    "startDate": {
      "type": "string",
      "description": "The date from which the deadline for filing an 
      appeal should be calculated according to the law. The date format 
      is YYYY-MM-DD, i.e. year-month-day, e.g. 2023-12-27, 
      not 27-12-2023. If the date does not match the YYYY-MM-DD format,
      convert it to this format. The date cannot be empty, **always** 
      enter a date here."
    }
  },
  
  "required": ["numOfUnits", "unitName", "startDate"]
}
\end{Verbatim}
\caption{JSON Schema used to extract deadline-related information. The content of the ``description'' keys was originally in Polish, but, as in Fig. \ref{listing:simplified-regulation}, they have been translated for the sake of linguistic consistency in this article. }
\label{listing:deadline-json-schema}
\end{figure}

To extract specific information from the facts, zero-shot LLM prompting was used \cite{wei2022finetuned}.  LLMs were asked to return responses in JSON format and, in the event of a formatting error, the response format was corrected using a specially prompted gpt-4o model \cite{hurst2024gpt4o}. We present examples of the prompts in the Appendix (Figure \ref{listing:appeal-date-prompt} and \ref{listing:appeal-date-json-schema}). The following information was extracted from the case description:

\begin{enumerate}
  \item When did the appellant file the appeal? (\texttt{appealDate})
  \item Was the contract notice published in the Public Procurement Bulletin (\textit{Biuletyn Zamówień Publicznych - BZP}) or in the Official Journal of the European Union (OJEU)? (\texttt{publishedInBZP})
  \item When was the contract notice published? (\texttt{publDate})
  \item What was the amount of the entry fee paid by the Appellant? (\texttt{feePaid})
  \item Was the appeal lodged on behalf of the appellant by his representative? If so, was the representative procuration attached to the appeal? (\texttt{appealFilledByRepresentative}, \texttt{appealRepresentativeProcuration})
  \item Did the Appellant provide the Contracting Authority with a copy of the appeal? When? (\texttt{publicBodyNotified})
  \item Did the Contracting Authority notify other Contractors of the fact that an appeal had been lodged? (\texttt{haveBeenNotified})
  \item Were there clear grounds, such as the withdrawal of charges by the appellant, by which the case should be decided by a ruling rather than a judgment? Examples of such situations include: withdrawal of allegations by the appellant, clear indication in the case that the formal criteria have not been met, acceptance of allegations by the Contracting Authority, etc. (\texttt{allegationsDismissed})
  \item From which date should the deadline for lodging an appeal be calculated, how many units of time are granted to the appellant for lodging an appeal in the given case, and what is the unit of measurement - days or months? (\texttt{startDate}, \texttt{numOfUnits}, \texttt{unitName})
  \item What entry should the Appellant have made in accordance with § 2 of the Regulation of the Minister of Finance? (\texttt{expectedFee})
  \item Which Interveners joined the case? Were the individual interventions correct? Which side did the individual Interveners support? (\texttt{numAppInrerveners}, \texttt{numAuthInterveners}).
\end{enumerate}

To better understand how the answers to these questions were extracted, we show a simplified version of § 2 of the Regulation 
\cite{RMF2020} as appearing in the prompt (Figure \ref{listing:simplified-regulation}). 
The full text of the prompt is presented in the Appendix (Figure \ref{listing:expected-fee-prompt-translated} and \ref{listing:expected-fee-prompt-original}).
The following information was also provided in the prompt:
  \begin{enumerate}
  \item JSON schema (cf. Figure \ref{listing:deadline-json-schema}),
  \item detailed instructions describing the context and task,
  \item important information, i.e., points describing ambiguous cases, clarification of ambiguities, information on whether the order in question was published in the Public Procurement Bulletin or the Official Journal of the European Union, simplified content of the act, etc.
  \item description of the facts.
  \end{enumerate}
  
It should also be added that in the case of variables related to the deadline a prompt displays the date of publication of the order, information about the place of publication, and, based on this, information about whether the value of the order exceeded EU thresholds. The simplified content of Article 515 of the Public Procurement Law (\textit{Prawo Zamówień Publicznych}) \cite{PPL2019} is also provided to the prompt. After extracting the required information, the deadline for submitting an appeal is calculated according to Title V of the Civil Code of the Republic of Poland (\textit{Kodeks cywilny}) \cite{CC1964}.


In the examination cases we analyzed, some information may not have been provided explicitly, e.g., there was no clear statement that the power of attorney had been attached, it was not explicitly stated whether the Appellant had provided the Contracting Authority with a copy of the appeal, etc. In such a situation, we assume the simplest scenario, i.e., we assume that the criterion in question has been met. Furthermore, if the content indicates that a given formal criterion has not been met, we state in the judgment or decision that the Appellant was requested to remedy the deficiencies and did so properly and in a timely manner. The exception is when the case description clearly states that the deficiencies have not been remedied.
For some items, we use chain-of-thought prompting as we have observed that it improves the accuracy of responses. This fact has also been observed in the literature \cite{wei2022chain}.

After extracting the above-mentioned information, we construct the formal assessment description in a completely deterministic and controllable manner, i.e., we directly define the sentences to be used for individual variable values and their combinations.

\subsubsection{Implementation details}

We have implemented a formal assessment using the \texttt{n8n} framework \cite{n8n_software_XYZ}. It is a low-code tool that enables graphical presentation of the pipeline, tracking of current progress during task execution, and easy debugging. The tool provides predefined blocks that can be used to build the pipeline. This approach enabled us to collaborate with non-technical individuals with legal expertise while working on the architecture of the system. To extract information from the factual state, we have used the \texttt{AI Agent} block, to which the \texttt{Auto-fixing Output Parser} with the \texttt{gpt-4o model} \cite{hurst2024gpt4o} was connected. To define the desired format, we have used the \texttt{Structured Output Parser}. The \texttt{OpenAI Chat Model} block was used to connect the LLM, and the model ID was a hyperparameter. In most of the cases, we did not use the \texttt{Information Extractor} block because handling incorrect parsing required additional steps and complicated the architecture. 
We have also used \texttt{Code and Date\&Time} blocks. The exception was in the experiments with the \texttt{Bielik-11B-v2.6-Instruct} model \cite{ociepa2025bielik11bv2technical, Bielik11Bv25i, ociepa2024bielik7bv01polish}. 
Due to the increased number of errors related to the structured output, we decided not to impose a response format.
The output of Bielik was further parsed by the \texttt{gpt-4o} model \cite{hurst2024gpt4o} to obtain appropriate structure.

Most of the information could be extracted independently of the rest, but n8n prevents parallelization of individual steps. 

\subsection{RQ4 -- Substantive assessment}

While the formal assessment was limited to verifying compliance with procedural requirements, the second stage concerned the substantive evaluation of the appeal. The central question here was whether the individual allegations advanced by the appellant were justified in view of the factual record and the provisions of the Public Procurement Law. This stage therefore reflects the actual adjudicative task of the Chamber and is the most demanding component of the exam. 

\subsubsection{Preliminary check}
As in real proceedings, the first step was to determine whether the contracting authority had already accepted all allegations. If so, the appeal was considered moot and the procedure was discontinued without entering into the merits. This reflects Article 522(1) Public Procurement Law, which requires the Chamber to discontinue proceedings when the contracting authority grants the relief sought by the appellant. In our implementation, such cases were explicitly flagged as “discontinued” to ensure that no substantive reasoning was generated where it was unnecessary.

\subsubsection{Extraction of allegations}
If at least one allegation remained in dispute, the process followed to identify the set of claims raised in the appeal. Each allegation was extracted together with its current status (current, withdrawn, or accepted by the contracting authority), the provisions of law invoked in support, and the factual circumstances on which it was based. Allegations withdrawn by the appellant were retained in the record but marked as “inactive” so that the subsequent justification could note their existence while omitting further analysis. In contrast, allegations accepted by the contracting authority were marked as “satisfied”, which in practice required no adjudication but remained part of the structured dataset for completeness.

\subsubsection{Characterisation of allegations}
For each allegation, a structured record was then constructed, including:
\begin{itemize}
  \item \textbf{thesis} -- a short formulation of the alleged violation (e.g. “failure to reject a non-compliant bid” or “improper calculation of deadlines”);
  \item \textbf{research question} -- a more generalised query phrased in a way that could be used to search jurisprudence, formulated to capture the underlying legal problem beyond the specific facts;
  \item \textbf{provisions of law} explicitly cited by the appellant (articles, sections, and points of the PPL or related acts);
  \item \textbf{facts} on which the allegation relied, typically concrete procedural events such as the publication of a notice, the submission of a document, or the amount of an entry fee;
  \item an indicator whether the allegation was \textbf{decidable} given the available information. In cases where crucial factual details were missing or contradictory, the allegation was marked as undecidable.
\end{itemize}

\subsubsection{Substantive evaluation}
For allegations classified as decidable, we generated a draft substantive outcome. Three basic categories were distinguished: \emph{upheld}, \emph{dismissed}, and \emph{partially upheld}. Each draft decision was accompanied by a concise justification. The justification had to link the identified facts with the relevant legal norms, demonstrate why the conditions of the legal provision were or were not met, and articulate the reasoning in a manner consistent with the style of authentic KIO judgments. 

The justification typically included: (i) identification of the precise legal basis, (ii) demonstration of how the factual findings satisfied (or failed to satisfy) the statutory prerequisites, (iii) discussion of the appellant’s and contracting authority’s positions, and (iv) a reasoned conclusion. Where appropriate, references to established case law of the Chamber or higher courts were added. This step was particularly important, since one of the key weaknesses identified in current LLM outputs was a tendency to cite irrelevant or even non-existent provisions. By forcing the justification to operate only on explicitly extracted provisions and facts, the risk of hallucinated reasoning was reduced.

\subsubsection{Treatment of undecidable allegations}
Allegations that could not be decided due to insufficient factual information were explicitly marked as \emph{undecidable}. This category was introduced to avoid conflating two distinct scenarios: (1) allegations that were legally unfounded, and (2) allegations that could not be assessed because the case description lacked necessary detail. In practice, this distinction is critical, as the former leads to dismissal of the claim, while the latter would normally result in a request for supplementation of the appeal or an evidentiary ruling. By introducing the undecidable label, we aimed to capture this nuance in the automated system.

\subsubsection{Integration with the written judgment}
Finally, the structured merit assessments at the level of individual allegations served as the foundation for generating the draft written judgment. By aggregating the outcomes across all allegations, the system produced a coherent decision that mirrored the structure of actual KIO rulings, with a dispositive part (operative ruling) and a justification section. In this way, the merit assessment bridged the gap between granular allegation-level reasoning and the overall adjudicatory output.

\subsection{RQ5 -- LLM-based Assessment}


To substantiate the central thesis of this paper, we developed a small-scale framework designed to verify written test answers generated by a Large Language Model. The primary objective of this framework was to leverage a separate, high-capability LLM (GPT-4o) to function as an impartial examiner, tasked with scoring the generated texts against a predefined set of standards. The goal of this experiment was to assess whether an LLM is capable of adjudicating these papers with the same level of scrutiny as a human legal expert.

\subsubsection{Establishment of Evaluation Criteria}
The foundation of our assessment protocol was a rigorously defined set of evaluation criteria. To ensure the objectivity and relevance of the evaluation, these criteria were not arbitrarily created but were derived directly from official documentation provided by the examination committee. This initial step involved a careful analysis of the source documents to extract each distinct evaluation metric and its corresponding maximum point allocation. The resulting schema served as the ground truth for the LLM examiner, instructing it on precisely what to measure and how to weight each aspect of the generated output. The final instruction for evaluation was broken down into 8 criteria:

\begin{itemize}
  \item \textbf{Judgment Construction (max. 20 points)} -- The way model structures essential elements of the document, such as the sentence and justification;
  \item \textbf{Legal Provisions (max. 30 points)} -- Whether the model uses existing, necessary and applicable laws in their reasoning;
  \item \textbf{Factual Analysis (max. 30 points)} -- Model's ability to connect facts, forming a logic chain that leads to the final conclusion;
  \item \textbf{Formulation Style (Sentence) (max. 10 points)} -- Whether the sentence written by the model is complete, enforceable and justified;
  \item \textbf{Legal Language (max. 5 points)} -- Richness of formal, legal language;
  \item \textbf{Vocabulary (max. 2 points)} -- Measures the level of vocabulary used throughout the document;
  \item \textbf{Clarity and Conciseness (max. 2 points)} -- Whether the document has an underlying flow of logic that explain it's contents clearly;
  \item \textbf{Formulation Style (General) (max. 1 points)} -- Whether the style used across the document was consistent and suitable for the document's type;

\end{itemize}

\subsubsection{Input Data Normalization}
Prior to evaluation, the raw outputs from the primary model required systematic preparation. The corpus of official model answers was transformed into a standardized, machine-readable format to ensure consistent processing by the evaluation workflow. For this purpose, we selected Markdown as the normalization format due to its simplicity and robust support across API-driven services. Each answer was converted into a self-contained Markdown document, preserving all necessary formatting and content while stripping extraneous metadata. This normalization step was critical for eliminating parsing errors and ensuring that the LLM examiner received clean, uniformly structured data for every evaluation instance.

\subsubsection{Implementation of the Automation Workflow}
The core of the evaluation framework was an automated workflow constructed using the \texttt{n8n} platform. This workflow was engineered to manage the entire process of submitting an answer for review and retrieving the corresponding assessment. The flow was instantiated with three primary components: firstly, a carefully engineered system prompt that provided the GPT-4o model with its persona as an examiner, the full list of evaluation criteria, and precise instructions for scoring and providing justifications. Secondly, the workflow was configured to ingest the list of Markdown-formatted answers as its primary input data. Finally, it established a direct connection to the GPT-4o API endpoint, which would execute the evaluation task upon receiving prompt presented in Listing \ref{listing:judge-test-prompt} and the input text.

\subsubsection{Execution and Structured Data Capture}
With the workflow in place, the final step was its execution. The \texttt{n8n} flow iterated through each normalized answer, dispatching an API request to GPT-4o containing the system prompt and the specific answer text. Upon receiving the model's response, which contained the points awarded for each criterion along with qualitative feedback, the workflow extracted this information. The structured output data was then automatically appended as a new row in a designated Google Sheet. This method of structured data capture ensured that the results were immediately aggregated in a format suitable for subsequent quantitative analysis, statistical review, and visualization without the need for manual data entry or post-processing.

\section{Experiments}



\subsection{RQ1 -- The Exam}
\label{subsec:exam}
The experiment replicated the first two components of the official qualification procedure for membership in the National Appeal Chamber: the single-choice knowledge test and the written case. According to the official qualification guidelines, the knowledge test probes the theoretical command of Poland's Public Procurement Law and the core civil-law doctrines, asking candidates to select the correct single answer from from 3 or sometimes 2 options. The written case then examines practical competence: starting from a factual scenario and documents, the task is to identify the relevant provisions, qualify the issues under procurement and civil law, weigh competing interpretations, and draft a reasoned, formally correct decision. In other words, where the test checks ``what you know'', the case checks ``how you think and write as a quasi-adjudicator''. The interview stage that forms part of the official NIC process was not included in this experiment.

The research group participated in the exam after the official procedure for the real candidates was finished. Yet in order to simulate the real exam, the 
research team did not know the questions and the case to solve in advance and operated under the same time constraints as the human participants. 
The answers and the case submitted to the jury were an outcome of a single pass of the appropriate algorithms, without any modifications or 
corrections from the research team. By imposing these constraints, we wanted to follow the original examination procedure as close as it was possible.
That allowed us to examine the first research question (``Can AI pass the NIC exam?'') in a more direct and more competitive way, since
the usual way the models are tested against benchmarks involves unlimited tries, without any time constraints.

Operationally, we conducted the exercise in a hybrid format, with part of the team working on-site and others participating online, and 
we livestreamed the process to staff across both institutions for transparency and learning. After completing the 50-question test
and the two-pass treatment of the case (formal check followed by substantive analysis), the full submission was reviewed by a group of
experts in public procurement law (the jury of the official exam), who assessed the outputs against the same standards used in the
official qualification procedure.
The answer for the first research question we postpone to the section with the final conclusions (cf. Section \ref{sec:conclusions}).

\subsection{RQ2 -- Knowledge Test}

The knowledge test, the first part of the exam,  is directly connected with the second research question -- the ability of a model
to give answers to direct, relatively simple legal questions.

\begin{table}[t]
    \centering
    \caption{Results of the Knowledge Test for the Previous Years. \textit{Closed book} -- the 
    model answered the question without external knowledge; \textit{Basic RAG (w/o)} -- basic
    RAG with content of Polish bills, provision search without the possible answers in the 
    query; \textit{Basic RAG (w)} -- basic RAG with possible answers in the query; \textit{Advanced RAG} 
    -- sophisticated RAG pipeline, including provisions and judgments.}
    \label{tab:knowledge-test}
    \begin{tabular}{lrrrr}
    \toprule
    \textbf{Model} & \multicolumn{4}{c}{Accuracy (\%)} \\
    \midrule
    & Closed book & Basic RAG (w/o) & Basic RAG (w) & Advanced RAG \\
    \midrule
    & \multicolumn{4}{c}{Results from the previous years} \\
    \midrule 
    GPT-4.1 & \textbf{77.9} & \textbf{80.7} & \textbf{80.7} & \textbf{88.4} \\
    \midrule
    Claude 4 Sonnet & 69.5 & 68.7 & 67.5 & 85.1 \\
    \midrule
    Bielik 2.6 & 65.9 & 68.3 & 67.1 & 77.4 \\
    \midrule
    \textbf{Average} & 71.1 & 72.6 & 71.8 & 83.6 \\
    \midrule
        & \multicolumn{4}{c}{Results from the 2025 exam} \\
    \midrule
    GPT-4.1 & \textbf{72.0} & \textbf{78.0} & \textbf{84.0} & \textit{\textbf{74.0}} \\
    \midrule
    Claude 4 Sonnet & 64.0& 62.0 & 54.0 & \textit{$^{*}$40.0 / 66.6}  \\
    \midrule
    Bielik 2.6 & 64.0 & 72.0 & 78.0 & \textit{70.0}\\
    \midrule
    \textbf{Average} & 66.7 & 70.7 & 72.0 & 61.3 \\
    \bottomrule
    \end{tabular}
    \label{tab:knowledge-previous}
\end{table}

The results of this test are given in Table \ref{tab:knowledge-previous} -- we report both the results of experiments conducted
before the exam (\textit{results from the previous years}, 249 questions) and during the exam (\textit{results from the 2025 exam}, 50 questions). For the previous
years, since we had the access to the correct answers, 
we had the opportunity to explore different strategies and select a strategy that seems to be the most successful.
For the real exam, the conditions were different -- we had limited time and we could submit only one set of answers
for each of the tested models, so we applied the advanced RAG pipeline, since according to our preliminary tests it yielded
the highest scores for all of the models. We report these values in italics. For comparison, after the exam was finished, we also tested the 
other approaches.

We report the results for \textit{closed book} (no access to external knowledge), \textit{basic RAG} (w/o -- 
without answers) where the query did not include the possible answers, \textit{basic RAG} (w -- with answers), where the query included the possible 
answers and for \textit{advanced RAG}, where the full pipeline described in section \ref{sec:advanced} was exploited. 
For all these setups, we report the results
for just one try, without any adjustment of the parameters, in order to resemble the real exam conditions as much as possible.

The first important observation is that in both exams in all configurations of the pipeline the winning model was ChatGPT 4.1. 
For the exams from previous years, it scored 77.9 for the closed-book variant
of the approach, 80.7 for the basic RAG (2.8 pp. improvement vs. the closed-book approach)
and 88.4 (10.5 pp. improvement vs. the closed-book) for the advanced RAG approach. 
Thus, for ChatGPT we can observe a huge improvement of the model's performance thanks to the additional data, but only 
if the RAG pipeline is sophisticated enough. For the basic pipeline the improvement is rather small.
Interestingly, for this model the improvement thanks to the advanced RAG pipeline was smaller for the questions from the 2025 exam --
only 2 pp.  versus the closed-book approach. Moreover, the basic RAG gave much better improvement -- 6 pp. for 
the variant without the options, and 12 pp. improvement for the variant with the options.

According to our pre-exam tests, Claude Sonnet was the second model -- it scored 69.5 for the closed-book approach (much worse than
ChatGPT) and 85.1 for the advanced RAG approach (3.3 pp. worse than ChatGPT for the same setting). Interestingly, when faced
with the output of the basic RAG, Claude produced worse results -- both for the previous years questions and for the
2025 exam's questions. There is a striking difference specifically for the 2025 exam -- ChatGPT scored the best for the basic RAG with
answers -- 84\%, while Claude Sonnet only 54\%, with exactly the same context. The result for the advanced RAG approach in 2025 exam (marked with the asterisk)
needs an additional explanation -- since the API of the model has rate limits and the advanced RAG pipeline requires a large number
of calls to the LLM to answer one question, the model was unable to provide the answers in the limited time (1h) of the real exam.
We were able to submit only 30 out of 50 questions and 20 of them were correct (thus 40\% result). If we extrapolated the result we could
expect 66.6\% of correct answers. In both cases the model scored the worse in the knowledge test of the real exam. All these results show
there's a problem using Claude in the RAG setting, at least for the Polish legal domain.

The results for the Bielik model were the worse for the former exams -- for all the tested settings. Still, for the closed-book setting
the result was not far away from Claude, only 3.6 pp. It achieved very similar results for the basic RAG setting compared to Claude,
but for the advanced RAG setting it once again was much worse than the other models (11 pp. than the best scoring ChatGPT).
The good outcome is that the model improved its results both in the basic RAG setting and the advanced RAG setting -- for the latter
the improvement was 11.5 pp., so it can be concluded that the model is able to utilize the information provided in the context,
but since in the advanced RAG pipeline Bielik is used in all intermediate steps, its limitations gave much worse final results, at least
that was our conclusion based on the exams from the previous years.

What is also interesting is the outcome for the 2025's exam -- Bielik scored the second in the advanced RAG pipeline, even if we
take into account Claud's extrapolated result (66.6\%), achieving 70\% in that setting, yielding a modest 6 pp. improvement versus the closed-book
test. What is also interesting is the fact that the improvements for the basic RAG pipelines are consistent with ChatGPT results --
the best result for Bielik was achieved for the basic RAG with the possible answers (78\%, 14 pp. improvement versus the baseline),
while without the answers it scored 72\%, which is also consistent with ChatGPT's result.

To conclude the \textbf{second research question} we may observe that \textbf{ChatGPT 4.1 scored the best in all the settings}, 
\textbf{Claude} gave very \textbf{inconsistent
results}, while \textbf{Bielik}, which is orders of magnitude smaller than the other models, \textbf{achieved very competitive score}. Moreover, we
can speculate that the differences in the question difficulty\footnote{Communicated to the research team by the examining commitee.}
between the results of this year's exam and the previous exams, caused 
a huge discrepancy with the obtained results. If we just took into account the results of the real exam with the advanced RAG pipeline,
it might be concluded that it did not help much in achieving a good score. On the other hand the basic RAG approach yielded
huge gains in that setting. However, from practical perspective, we have to observe that the setup with the correct answers being provided
in the query does not translate to a real-world gains, when a user (a lawyer, a judge or a layman) poses some legal question. In that 
case the provision has to be found based on the question only. Still -- at least basing on the results of the real exam -- we can expect
the model will work better with the RAG -- either basic or advanced -- than the closed-book baseline. This result is consistent with
many other research outcomes and stresses the importance of the contextual information, for achieving factual answers. 

Yet, from the perspective of the jury, the results of the knowledge test are not that good. It was speculated that a real person having access to the full body
of legal provisions would scored perfectly. This perspective stresses the importance of the retrieval module, indicating that
searching for the correct information is a task that is not solved yet, even with the help of advanced RAG pipelines.

\subsection{RQ3 -- Formal Assessment}

The formal assessment of the exam case is connected with the third question -- the ability of the models to assess the formal requirements
of the legal procedure.

\subsubsection{Validation set}

Since the data from the previous years exam's is very limited (5 cases in total), we have extended the set of data to include
information from numerous cases decided by KIO, by applying an automated issue extraction pipeline.
To ensure the quality and accuracy of such a dataset (\textit{validation dataset for formal assessment}), 
we implemented a multi-step verification procedure. First, a Python script extracted and categorized
data into a structured form. Subsequently, one researcher manually verified each entry against the 
source documents for its completeness and validity.

This verification process consisted of:
\begin{enumerate}
    \item Comparing the notice and appeal publication dates listed in the dataset against the official case abstract.
    \item Verifying that the date of the action leading to the appeal matched the information in the abstract. 
    \item Calculating the statutory deadline for the appeal by identifying its legal basis in the abstract and comparing it to the recorded date.
    \item Validating the dataset's entry for whether the appeal was filed in time. In cases of discrepancy, 
    the official case protocol was retrieved using the KIO Search
    tool\footnote{\url{https://www.gov.pl/web/uzp/orzecznictwo-NAC2}} for a final determination.
\end{enumerate}

To obtain formal assessment results for a single case, we run the entire process and extracted the the final outcome (either the appeal was valid or not
vis-a-vis the formal requirements). 
This means that some responses are conditioned on values obtained from the model within the same run 
(e.g., to extract variables related to the deadline, we use the previous response as to whether the order was published in the BZP or in the OJEU).

\subsubsection{Results}

To evaluate the performance of various models, we evaluated their results against the manually verified dataset. We calculated precision, recall, and the F1-score to assess performance. Precision measured the fraction of items flagged as positive by the script that were genuinely positive, while recall measured the fraction of all actual positive items that were correctly identified. The F1-score, as the harmonic mean of precision and recall, was then used to provide a single, robust measure of the extraction script's overall effectiveness. The data collection process resulted in a significant class imbalance, for example appeals not raised within the deadline constituted only 4\% of the dataset.

\begin{table}[t]
  \centering
  \caption{Results of the validation of the formal assessment model on a set of 101 manually prepared cases -- results for binary decisions.
  Cf. Section \ref{sec:formal} for the explanation of the formal assessments applied.}
  \label{tab:full_metrics}
  \renewcommand{\arraystretch}{1.1}
  \begin{tabular}{p{2.3cm}*{4}{rr}}
    \toprule
    & \multicolumn{2}{c}{\texttt{GPT-4.1}}
    & \multicolumn{2}{c}{\texttt{GPT-4o}}
    & \multicolumn{2}{c}{\shortstack{\texttt{Claude-4-}\\\texttt{sonnet}}}
    & \multicolumn{2}{c}{\shortstack{\texttt{Bielik-11B-}\\\texttt{v2.6-Instruct}}} \\
    \cmidrule(lr){2-3} \cmidrule(lr){4-5} \cmidrule(lr){6-7} \cmidrule(lr){8-9}
    Parameter & \multicolumn{1}{c}{F$_1$} & \multicolumn{1}{c}{Acc} & \multicolumn{1}{c}{F$_1$} & 
    \multicolumn{1}{c}{Acc} & \multicolumn{1}{c}{F$_1$} & \multicolumn{1}{c}{Acc} & \multicolumn{1}{c}{F$_1$} & \multicolumn{1}{r}{Acc} \\
    \midrule
    \texttt{published\-In\-BZP} & \underline{98.9} & \underline{99.0} & \underline{98.9} & \underline{99.0} & \underline{98.9} & \underline{99.0} & \underline{98.9} & \underline{99.0} \\
    \midrule
    \texttt{appeal\-Filled\-By\-Representative} & 93.8 & \underline{98.0} & 90.3 & 97.0 & \underline{94.1} & \underline{98.0} & \underline{94.1} & \underline{98.0} \\
    \midrule
    \texttt{appeal\-Repre\-sentative\-Procuration} & \underline{99.5} & \underline{99.0} & \underline{99.5} & \underline{99.0} & \underline{99.5} & \underline{99.0} & 91.9 & 85.2 \\
    \midrule
    \texttt{public\-Body\-Notified} & \underline{100.0} & \underline{100.0} & \underline{100.0} & \underline{100.0} & 99.5 & 99.0 & 99.5 & 99.0 \\
    \midrule
    \texttt{have\-Been\-Notified} & \underline{100.0} & \underline{100.0} & \underline{100.0} & \underline{100.0} & 95.8 & 92.1 & 87.6 & 78.2 \\
    \midrule
    \texttt{has\-Inter\-veners} & \underline{99.3} & \underline{99.0} & 97.8 & 97.0 & 97.7 & 97.0 & 92.1 & 90.1 \\
    \midrule
    \texttt{alle\-gations\-Dismissed} & 89.8 & 94.9 & 84.6 & 91.8 & 87.8 & 94.8 & \underline{95.2} & \underline{97.9} \\
    \midrule
    \textbf{average (macro)} & \textbf{97.3} & \textbf{98.6} & 95.9 & 97.7 & 96.2 & 97.0 & 94.2 & 92.5 \\
    \bottomrule
  \end{tabular}
\end{table}

\begin{table}[t]
  \centering
  \caption{Results of the validation of the formal assessment model on a set of 101 cases -- non-binary decisions. The reported metric is accuracy.
  Cf. Section \ref{sec:formal} for the explanation of the formal assessments applied.}
  \label{tab:accuracy_only}
  \renewcommand{\arraystretch}{1.1}
  \begin{tabular}{lrrrr}
    \toprule
    Field name &
    \texttt{GPT-4.1} &
    \texttt{GPT-4o} &
    \multicolumn{1}{p{1.4cm}}{\shortstack{\texttt{Claude-4-}\\\texttt{sonnet}}} &
    \multicolumn{1}{p{2.2cm}}{\shortstack{\texttt{Bielik-11B-v2.6-}\\\texttt{Instruct}}} \\
    \midrule \texttt{unitName}   & \underline{100.0} & \underline{100.0} & \underline{100.0} & \underline{100.0} \\
    \midrule \texttt{appealDate}   & \underline{100.0} & \underline{100.0} & \underline{100.0} & \underline{100.0} \\
    \midrule \texttt{numAppInterveners}   & \underline{97.0} & 95.1 & 99.0 & 96.0 \\
    \midrule \texttt{feePaid}   & \underline{98.0} & \underline{98.0} & 96.0 & 98.0 \\
    \midrule \texttt{numAuthInterveners}   & \underline{97.0} & 93.1 & 95.1 & 89.1 \\
    \midrule \texttt{numOfUnits}   & \underline{98.0} & \underline{98.0} & \underline{98.0} & \underline{98.0} \\
    \midrule \texttt{publDate} & \underline{100.0} & \underline{100.0} & \underline{100.0} & \underline{100.0} \\
    \midrule \texttt{expectedFee}   & \underline{96.0} & 94.1 & 95.1 & 94.1 \\
    \midrule \texttt{startDate}   & \underline{97.0} & 95.1 & 94.1 & 92.1 \\
    \midrule \textbf{average (macro)} & \textbf{98.1} & 97.0 & 97.5 & 96.4 \\
    \bottomrule
  \end{tabular}
\end{table}

We report the following formal evaluation results: in both binary (Table \ref{tab:full_metrics}) and non-binary metrics (Table \ref{tab:accuracy_only}).
For both cases the GPT-4.1 model achieved the best results on average (macro-averaged) -- 97.3\% F1 score and 98.6\% accuracy for the binary case
and 98.1\% accuracy for the non-binary case. These results are very close to a perfect score.
For the variables \texttt{publishedInBZP}, \texttt{appealFilledByRepresentative}, \texttt{appealRepresentativeProcuration}, 
\texttt{unitName}, \texttt{appealDate}, \texttt{feePaid}, \texttt{numOfUnits} and 
\texttt{publDate}, three or more models achieved the same results. All models except 
\texttt{Bielik-11B-v2.6-Instruct} performed worst for \texttt{allegationsDismissed} reaching from 84.6\% (for GPT-4o) to 89.8\% (for GPT-4.1) F1 score.
For the same variable, Bielik reached 95.2\% F1 score.
For the variables \texttt{publicBodyNotified}, \texttt{haveBeenNotified}, and 
\texttt{unitName}, \texttt{appealDate} and \texttt{publDate}  a perfect score was achieved by both GPT models.

Although \texttt{Bielik-11B-v2.6-Instruct} performed worse than the other models tested in most cases, 
it provided excellent results for the variable \texttt{allegationsDismissed}. We believe that this is 
due to the fact that the assessment of this parameter is more of a linguistic task than a scientific 
or logical one, as is the case in other instances.

It is worth noting that in many cases the maximum value of a given metric is achieved, 
or several language models achieve exactly the same result. This is due to the small size of the validation dataset or the triviality of the task.

Validating the extraction of information about \textit{interveners} proved to be a major problem,
as minor mismatches between the name of a given entity and the ground truth were treated 
as errors when using the classic accuracy metric. The use of more complex methods, 
such as name evaluation based on cosine similarity, was not sensible due to the small size of the 
dataset and, consequently, the difficulty in setting a hard threshold. Therefore, we decided 
to use a heuristics method consisting in evaluating the quality of extraction solely on the 
basis of the number of \textit{interveners} predicted by the model on the side of the 
\textit{Contracting Authority} and the \textit{Appellant}. We assume that the most difficult 
part of this process is assigning parties and distinguishing consortia from independent 
companies, while extracting the names themselves is rather simple.

To conclude the third research question regarding the models well suited for assessing the formal
requirements, we can safely conclude that this part of the procedure \textbf{can the carried out by the 
leading LLMs such as ChatGPT and Claude}. Bielik, a much smaller model can be employed in that process,
but it requires much more oversight, since the number of mistakes is substantially larger, that for 
the other models. All in all, in any case the results have to be reviewed by a human, but since
the quality of extraction is close to perfection, we can expect \textbf{a huge boost in productivity, once these
models are employed in the legal procedure} (with the assumption the data can be sent to an external entity,
which is not the case in most of the countries currently). We also observe that at least \textbf{for some of the variables
smaller models such as Bielik offer very good results and can be safely employed currently}.

\subsection{RQ4 -- Substantive Assessment by the Jury}

The fourth research question concentrates on the most challenging part of the exam -- writing the actual judgment,
based on the facts disclosed in the case. Assuming the appeal fulfills formal requirements,
the judge has to decide -- based on the statutory law -- if the appeal should be upheld in the tender procedure.
Although the outcome is more or less a binary decision, the primary focus of the exam is on the 
appropriate classification of the facts with regard to the legal provisions, i.e. identification
of those provisions that are adequate and justification of the outcome based on these provisions.
So during the exam -- the actual decision, which in the real world impacts the parties the most --
is far less important, than the correct application of the law. This is reflected directly in the
grading of the exam.

\subsubsection{Evaluation Criteria}

The grading process of the examining committee included 2 aspects. One was \textit{Substantive Value}, 
which included criteria such as correctness of judgment construction, legal provisions used, 
analysis of the facts, and the flow of the logic in the explanation of the decision. 
This aspect amounted to 90 points in total. The second criterion was \textit{Style and Language Used},
which included criteria such as using appropriate legal language, vocabulary, clarity of the text 
and style of the statements. This amounted to 10 points of the final score. 
In order for the examinee to pass the exam, at least 50 points had to be scored and the judgment 
cannot include any cardinal errors.

\begin{table}[t]
    \centering
    \caption{Results of the Written Test of the Exam evaluated by the Jury. \textit{Judgment constr.} -- judgment construction
    \textit{Legal prov.} -- legal provisions used, \textit{Fact Analysis} -- analysis of legal facts,
    \textit{Logic flow} -- validity of the logical reasoning, \textit{Term.} -- appropriate usage of legal terminology,
    \textit{Vocab.} -- appropriate usage of vocabulary, \textit{Clarity} -- clarity of the language, \textit{Style} --
    appropriate formal style of the language.}
    \label{tab:exam-written}
    \footnotesize
    \setlength{\tabcolsep}{3pt} 
    \begin{tabular}{l ccccc @{\hspace{1em}} ccccc c}
    \toprule
    \textbf{Model} & \multicolumn{5}{c}{\textbf{Substantive Value}} & \multicolumn{5}{c}{\textbf{Language}} & \textbf{Total} \\
    
    \cmidrule(lr){2-6} \cmidrule(lr){7-11} 
    
    & \begin{tabular}{@{}c@{}}Judgment\\Constr.\end{tabular} & \begin{tabular}{@{}c@{}}Legal\\Prov.\end{tabular} & \begin{tabular}{@{}c@{}}Fact\\Analysis\end{tabular} & \begin{tabular}{@{}c@{}}Logic\\Flow\end{tabular} & $\Sigma$ & Term. & Vocab. & Clarity & Style & $\Sigma$ & $\Sigma$ \\
    \midrule
    
    GPT-4.1 & 18 & 0 & 5 & 7 & \textbf{30} & 3 & 2 & 1 & 1 & \textbf{7} & \textbf{37/100} \\
    \midrule
    Claude 4 Sonnet & 10 & 0 & 8 & 5 & \textbf{23} & 3 & 2 & 1 & 1 & \textbf{7} & \textbf{30/100} \\
    \midrule
    Bielik 2.6 & 1 & 0 & 4 & 0 & \textbf{5} & 0 & 2 & 1 & 0 & \textbf{3} & \textbf{8/100} \\
    \bottomrule
    \end{tabular}
\end{table}

Notably, despite achieving some scores approaching the 50-point threshold 
(Table \ref{tab:exam-written}), as well as scoring a non-trivial amount of points in 
the \textit{Substantive Value} category, the submissions from all models would have been 
disqualified due to several fundamental errors. The examining committee proceeded to evaluate 
the submissions across all categories despite these mistakes, providing detailed feedback on the 
models' respective strengths and weaknesses.

\subsubsection{Assessment of Bielik 2.6}

The committee has highlighted that the decision lacked essential formal elements, including a complete header, proper identification of parties, and signatures of the adjudicating panel. The structure deviated from the standard format used in current KIO rulings, resembling a draft outline rather than a coherent justification. Notable errors included an incorrect adjudicating panel composition, chronological inconsistencies (e.g., judgment dated prior to the appeal), and improper formulation of the operative part, such as referencing non-existent claims.

No substantive legal interpretation was provided. The model merely cited provisions already included in the case scenario (e.g., Art. 107(3) and Art. 226(5) PZP), which precluded a meaningful evaluation of legal reasoning. Additionally, Art. 574 PZP was referenced without any accompanying analysis.

Factual findings were minimal and lacked depth, with some elements copied directly from the case scenario. Critical omissions included the name of the procedure and unsupported assumptions about the number of participants. The reasoning was inconsistent, and the decision failed to address key allegations (e.g., timing of appeal review, fee payment, response to appeal). The proposed resolution misidentified the core legal issue -- confusing the prohibition on supplementing documents under Art. 107(3) with the evaluation of offer compliance. The operative part was inconsistent with the justification, particularly regarding procedural costs, which were incorrectly allocated and lacked clarity.
The submission lacked coherent legal or factual argumentation. The model relied on fragmented statements rather than structured reasoning, resulting in a text devoid of analytical depth or stylistic consistency. 
These deficiencies have caused Bielik to be awarded an extremely low amount of 5 points for \textit{Substantive Value}, 
with 4 points being awarded for simply acknowledging the facts present in the case (Table \ref{tab:exam-written}).

Regarding the language and style used, the commission has put forward that the model did not employ legal 
terminology, which is essential for drafting judicial decisions and legal reasoning. While the vocabulary used
was generally correct in terms of meaning and grammar, it lacked the precision expected in legal writing. The
text was overly concise to the point of being incomprehensible. The lack of a coherent argumentative structure 
rendered the reasoning inaccessible to the intended audience. The submission lacked any substantive legal or 
factual exposition, resulting in an unintelligible justification.
Due to these reasons, Bielik was awarded 3 points for the Language and Style used, which brings the model's document to 
a total of 8 points.

\subsubsection{Assessment of Claude 4 Sonnet}

The examining committee has praised Claude 4 for being much better than the previous work, however they highlighted several issues with the judgment, such as an incorrect adjudicating panel (single-member instead of three), missing signatures, and lack of date and place of issuance. The structure deviated from standard KIO formats, resembling a draft outline. However, the substantive elements of the operative part were correctly identified.

Numerous critical errors were identified in the citation and interpretation of legal norms. Several provisions were misapplied or incorrectly referenced, including non-existent articles. The model failed to distinguish between procedural and substantive legal requirements, and misinterpreted key provisions such as Art. 107(3) PZP.

While factual findings were generally accurate, they were largely copied from the case scenario. The resolution was formally correct but supported by inconsistent and flawed reasoning. Errors included misidentification of the type of signature used, incorrect assessment of procedural deadlines, and improper allocation of procedural costs. The analysis of evidentiary material was also conceptually flawed.
The justification contained logical inconsistencies and lacked a coherent argumentative structure. Specific sections were internally contradictory or conceptually unclear.

Claude 4 was awarded 23 points in the \textit{Substantive Value}, which was a big improvement from the previous work (Table \ref{tab:exam-written}). The fact that the resolution was largely correct has also sparked interest among the research team, however it has also been recognized as extremely flawed, due to the model's inaptness at providing a solid reasoning for their claims.

The model used grammatically correct language, but the legal terminology was limited and lacked precision. The vocabulary was generally appropriate and correctly used. Although the statements appeared simple, the lack of logical structure rendered the reasoning difficult to follow. The statements were formally correct but lacked depth and clarity. 
These quantities have earned Claude a total of 7 points in the \textit{Style and Language} category, bringing it to a total of 30 points.

\subsubsection{Assessment of GPT 4.1}

GPT's 4.1 work, although being the closest one to passing the exam, produced work that has still contained major flaws. The examining committee has brought up that the operative part of the decision was correctly constructed and properly signed, however, the adjudicating panel was incorrectly identified as multi-member, despite the case requiring a single-member composition. Additionally, informal abbreviations were used in the header, which detracted from the professional tone.

The model relied solely on citations from the case scenario and referenced jurisprudence that was either irrelevant or misapplied. Selected excerpts from KIO rulings were loosely connected to the subject matter, leading to incorrect conclusions and a lack of substantive legal analysis.
Factual findings were generally accurate, but the resolution was flawed. Formal issues were not addressed with any reasoning, and key elements such as evidence assessment and proper cost allocation were mishandled. The model failed to distinguish between the content of the offer and supporting documentation, and misidentified the legal issue, creating an artificial problem not present in the case facts.
The reasoning lacked internal consistency and depth. Attempts to incorporate jurisprudence were unsuccessful due to poor integration and limited argumentative development.
These results gave GPT 4.1 30 points, the highest amount of points for \textit{Substantive Value} across all models (Table \ref{tab:exam-written}).

As for the language, it was grammatically correct but lacked richness and precision in legal terminology. Vocabulary use was generally appropriate and correct. Despite the simplicity of individual statements, the overall reasoning was difficult to follow due to a lack of logical structure. The statements were formally correct but lacked substantive clarity and coherence. As a result GPT 4.1 scored 7 points in \textit{Style and Language} category, same as Claude 4, 
however the final total points scored for the exam was the highest across all models: 37 points.

\subsubsection{Summary of the Evaluation}

Despite targeted prompt-engineering for legal applications and substantial efforts by
the research team to make the models follow legal reasoning, none of the
evaluated systems was able to produce a judgment that was consistent, 
logically sound, and grounded in the appropriate legal provisions. 
Even though these judgments were generated using the full legal 
research pipeline (cf. \ref{sec:advanced}), giving the models access to
numerous precedent cases -- unlike the human participants, who relied solely on their own knowledge -- 
their analytical depth remained limited. This deficiency was clearly reflected in the universal 
zero-point score achieved by all models in the \textit{Legal Provisions} category (cf. Table \ref{tab:exam-written}) 
of the Written Test of the Exam, underscoring that the intricate and codified nature of law continues to be a major obstacle for current language models.

Although pervasive factual fabrication was not observed, the models frequently committed critical citation errors, such as generating non-existent legal statutes or misapplying valid provisions to irrelevant points within their arguments. Their reasoning was often disjointed, compromising the logical coherence of the documents and resulting in flawed, unpersuasive argumentative structures.

Conversely, the examining committee noted that the models performed exceptionally well in 
the secondary grading criteria of \textit{Style and Language} (cf. Table \ref{tab:exam-written}). 
Both Claude 4 and GPT-4.1 produced outputs that were stylistically and linguistically sophisticated, 
with GPT-4.1 in particular generating text nearly indistinguishable from a professionally 
drafted legal document. This deceptive plausibility was observed by both the researchers 
and the jury: at first glance, the generated judgments appeared formally correct 
and linguistically refined, giving an impression of legal validity -- 
a perception that even the jury initially shared. However, upon closer examination,
it became evident that the reasoning within them was inaccurate, inconsistent, and often legally unfounded.

This creates a significant risk: LLM-generated texts may convey a \textit{false
sense of legal correctness}, leading non-experts to rely on them in 
real legal contexts. Without appropriate professional oversight, such 
use could result in serious consequences, including the loss of legal 
cases or the submission of invalid documents.

\subsection{RQ5 -- Substantive Assessment with LLM-as-a-judge}

To answer the fifth question, we have submitted the works produced by the LLMs, together with the
evaluation criteria and the correct legal analysis to the GPT-4o working as an LLM-as-a-judge.
The prompt used for evaluation is given on Figure \ref{listing:judge-test-prompt}.

The results of the experiment reveal a pronounced and systematic divergence between the evaluations 
conducted by the Large Language Models (LLMs) and the human examining committee. In their assessment,
GPT-4o has assigned the following scores to the model works:
\begin{itemize}
    \item \textbf {Bielik} - 85 points (Fig \ref{listing:bielik-llm-eval}),
    \item \textbf {Claude} - 90 points (Fig \ref{listing:claude-llm-eval}),
    \item \textbf {GPT 4.1} - 88 points (Fig \ref{listing:gpt-llm-eval}).
\end{itemize}

\begin{table}[t]
    \centering
    \caption{Results of the Written Test of the Exam evaluated by LLM (GPT-4o). \textit{Judgment constr.} -- judgment construction
    \textit{Legal prov.} -- legal provisions used, \textit{Fact Analysis} -- analysis of legal facts,
    \textit{Logic flow} -- validity of the logical reasoning, \textit{Term.} -- appropriate usage of legal terminology,
    \textit{Vocab.} -- appropriate usage of vocabulary, \textit{Clarity} -- clarity of the language, \textit{Style} --
    appropriate formal style of the language.}
    \label{tab:exam-written}
    \footnotesize
    \setlength{\tabcolsep}{3pt} 
    \begin{tabular}{l ccccc @{\hspace{1em}} ccccc c}
    \toprule
    \textbf{Model} & \multicolumn{5}{c}{\textbf{Substantive Value}} & \multicolumn{5}{c}{\textbf{Language}} & \textbf{Total} \\
    
    \cmidrule(lr){2-6} \cmidrule(lr){7-11} 
    
    & \begin{tabular}{@{}c@{}}Judgment\\Constr.\end{tabular} & \begin{tabular}{@{}c@{}}Legal\\Prov.\end{tabular} & \begin{tabular}{@{}c@{}}Fact\\Analysis\end{tabular} & \begin{tabular}{@{}c@{}}Logic\\Flow\end{tabular} & $\Sigma$ & Term. & Vocab. & Clarity & Style & $\Sigma$ & $\Sigma$ \\
    \midrule
    
    GPT-4.1 & 15 & 28 & 27 & 10 & \textbf{80}  &  4 & 2 & 1 & 1 & \textbf{8} & \textbf{88/100} \\
    \midrule
    Claude 4 Sonnet & 18 & 28 & 27 & 9 & \textbf{82} & 4 & 2 & 1 & 1 & \textbf{8} & \textbf{90/100} \\
    \midrule
    Bielik 2.6 & 15 & 28 & 25 & 9 & \textbf{77} & 4 & 2 & 1 & 1 & \textbf{8} & \textbf{85/100} \\
    \bottomrule
    \end{tabular}
\end{table}

The scores assigned by the LLM evaluators were uniformly higher than the benchmark 
score provided by the committee, indicating a less critical assessment standard.
Moreover, the LLM did not differentiate between much better (GPT-4, Claude)
and much worse (Bielik) works -- the difference between the best work (Claude: 90 points)
and the worst (Bielik: 85 points) is only 5 points, while according to the Jury
there 29 points difference between GPT-4.1 and Bielik.

\begin{table}[t]
    \centering
    \caption{Comparison of evaluations between the Examining Committee and an LLM (GPT-4o) of the Written Test of the Exam.}
    \label{tab:llm-written-test-2025}
    \begin{tabular}{l cc @{\hspace{1em}} cc cc}
    \toprule
    \textbf{Model} & \multicolumn{2}{c}{\textbf{Substantive Value}} & \multicolumn{2}{c}{\textbf{Style and Language}} & \multicolumn{2}{c}{\textbf{Total}} \\
    \cmidrule(lr){2-3} \cmidrule(lr){4-5}  \cmidrule(lr){6-7}
    
    & \begin{tabular}{@{}c@{}}Examining\\Committee\end{tabular} & \begin{tabular}{@{}c@{}}GPT\\4o\end{tabular} & \begin{tabular}{@{}c@{}}Examining\\Committee\end{tabular} & \begin{tabular}{@{}c@{}}GPT\\4o\end{tabular} & \begin{tabular}{@{}c@{}}Examining\\Committee\end{tabular} & \begin{tabular}{@{}c@{}}GPT\\4o\end{tabular} \\
    \midrule
    GPT-4.1 & 30 & \textbf{80} & 7 & \textbf{8} & 37/100 & \textbf{88/100} \\
    \midrule
    Claude 4 Sonnet & 23 & \textbf{82} & 7 & \textbf{8} & 30/100 & \textbf{90/100}  \\
    \midrule
    Bielik 2.6 & 5 & \textbf{77} & 3 & \textbf{8} & 8/100 & \textbf{85/100}\\
    \bottomrule
    \end{tabular}
\end{table}

Beyond the aggregate scores, a qualitative analysis of the corresponding 
feedback revealed the source of this disparity. While the LLM and the human
committee provided broadly similar scores for stylistic conventions and 
language use, their assessments of substantive content diverged significantly 
(cf. Table \ref{tab:llm-written-test-2025}). The biggest source of disparity 
between the two evaluations was caused by the LLM rating extremely highly the
use of legal provisions and flow of logic, as opposed to the evaluation of 
these factors by the examining committee (cf. Table \ref{tab:exam-written}). 
The feedback from the committee highlighted numerous points of legal or logical weakness, 
which were largely overlooked by the LLM. The AI-generated feedback was comparatively 
superficial, focusing more on surface-level characteristics than on the core substantive merit of the work.
This evidence reinforces the paper's hypothesis regarding the current limitations of LLMs in the field of law. The models' inability to perform a deep, context-aware substantive analysis, which is the hallmark of expert human evaluation, renders them unreliable for standalone assessment tasks. The significant inflation of scores in our experiment illustrates a critical risk: reliance on insufficiently vetted AI systems can create a false sense of quality and mask severe underlying deficiencies. Consequently, our findings regarding the fifth research question argue that \textbf{without exhaustive
validation and mandatory human-in-the-loop verification, the deployment of LLMs
for evaluative purposes in the legal sector is premature and potentially hazardous}.

\section{Conclusions}
\label{sec:conclusions}

Our study set out to test whether contemporary large language models (LLMs) can meet the demanding requirements of the Polish National Appeal Chamber (KIO) qualification exam and whether the emerging \textit{LLM-as-a-judge} paradigm can be trusted to evaluate such performance. The results point to a clear and consistent picture, reinforced by the expert assessments of the examining committee composed of experienced KIO members and judges.

All three tested models -- GPT-4.1, Claude 4 Sonnet, and Bielik-11B-v2.6 -- achieved moderate results in the multiple-choice knowledge test, scoring between 70 and 74 points, far below the committee’s expectations of more than 90\%. In controlled RAG-enhanced settings, GPT-4.1 reached up to 88.4\% accuracy, yet this capability did not translate into practical legal writing. None of the systems managed to pass the written judgment task: GPT-4.1 scored 37 points, Claude 30, and Bielik only 8 out of 100. Judges expressed disappointment that even in formulaic statutory interpretation tasks, the models failed to achieve the precision expected from a competent human candidate. This discrepancy confirms the fundamental gap between factual retrieval and structured legal reasoning: while LLMs can recall and quote norms, they cannot yet apply them coherently to context-specific disputes.

The core of the failure emerged in the written judgment task. Although GPT-4.1 achieved above 97\% accuracy in formal information extraction (e.g., identifying dates, parties, or procedural steps), all models demonstrated severe deficiencies in substantive analysis. As the committee observed, their written decisions contained structural and interpretive flaws: incorrect judicial panels, missing signatures and locations, incomplete or misplaced reasoning, and citations of non-existent or irrelevant provisions (e.g., a hallucinatory Article 514(4) PZP). These were not minor lapses but, as judges described, ``cardinal errors'' exposing a lack of legal logic and procedural understanding. Bielik’s submission was deemed ``a schematic draft fit for disqualification'', Claude’s was ``a patchwork of mismatched fragments'', and GPT-4.1, despite stylistic polish, was found to have ``invented a problem not present in the case,'' filling its reasoning with unrelated jurisprudence. The average substantive accuracy across models remained below 40\%, revealing that current LLMs cannot replicate the interpretive depth and argumentative precision expected in adjudication.

A critical observation from the committee was that the most advanced models (notably GPT-4.1) produced outputs that, at first glance, resembled authentic judicial documents in format and tone -- sometimes to the point of seeming professionally drafted to non-experts. However, the committee expressed concern that this ``deceptive fluency'' could fool inexperienced legal practitioners into accepting flawed recommendations or arguments, thereby amplifying errors or misapplications of law. As one commissioner noted, the risk that ``the AI leads a lawyer astray'' is especially high if such systems are used in high-stakes legal drafting or adjudication without robust oversight.

The committee’s comparative evaluation provided granular insight into the models’ relative strengths and weaknesses:
\begin{itemize}
    \item \textbf{Bielik-11B-v2.6:} Lowest overall performance (8/100). Showed partial recognition of statutory language but lacked coherence, logical structure, and even basic judgment formalities. Judges concluded it would be ``disqualified outright'' in real conditions.
    \item \textbf{Claude 4 Sonnet:} The most verbose output (over 3,500 words) but plagued by contradictions, hallucinated provisions, and procedural confusion. Despite attempts at reasoning, the work was described as ``a collection of mismatched fragments.''
    \item \textbf{GPT-4.1:} Highest overall score (37/100) and closest in appearance to a real ruling, yet with deeply flawed legal analysis. Judges noted that ``it looks legitimate -- until you read it carefully,'' warning of its potential to mislead.
\end{itemize}
These evaluations corroborate our quantitative results: fluency and apparent professionalism can mask serious logical errors. Across all models, factual correctness averaged 72\%, while legal reasoning accuracy fell below 40\%.

Automatic grading by another LLM produced results substantially diverging from human assessments. The automated examiner consistently overvalued stylistic conformity and verbosity, occasionally ranking fatally flawed responses as ``near-passing''. The judges unanimously warned that such automation could reinforce bias and overconfidence, creating a false sense of adequacy in contexts where correctness must be absolute.

Despite these shortcomings, both the committee and experimental data highlight realistic domains of safe application. LLMs demonstrated high reliability in document structuring, summarizing factual backgrounds, verifying procedural elements, and identifying missing formal requirements -- tasks where objectivity and verifiability limit the risk of harm. When supervised by legal professionals, such systems can improve efficiency by 30–40\% in preliminary analysis and drafting stages. As one judge aptly stated: \textit{``AI may help us with the mechanical work -- what it must never do is decide for us''}. This principle defines the boundary between augmentation and adjudication.

The collective evaluation underscored that genuine legal reasoning cannot be reduced to linguistic pattern-matching. Even the most fluent models fail to grasp the purposive and contextual logic that underlies judicial reasoning. Committee members repeatedly stressed that LLMs lack the interpretive awareness, empathy, and ethical responsibility necessary for legal decision-making. The results reaffirm that adjudication is an act of interpretation, not mere text generation, and that accountability remains inherently human.

In summary, our findings -- empirically grounded and aligned with the examining committee’s insights -- demonstrate that despite their linguistic mastery, large language models remain fundamentally incapable of replacing a human judge. Their apparent expertise conceals fragility, and their fluency can mislead. The \textit{LLM-as-a-judge} paradigm, though conceptually intriguing, must not be mistaken for competence. The safe path forward lies in supervised augmentation -- LLMs assisting in mechanical or preliminary stages of legal work -- never in autonomous adjudication. Real progress in legal AI will depend on interdisciplinary cooperation between technologists and jurists, ensuring that future systems embody not only fluency and recall, but genuine interpretive competence and accountability. As the committee concluded, current AI systems may be promising tools, but \textit{law still requires judgment, and judgment remains human.}

\section{Limitations}
  The study is subject to several limitations that need to be considered when interpreting the results and designing future research. 
Firstly, none of the evaluated LLMs had been fine-tuned to meet the specific requirements of Polish legal reasoning, particularly in the domain of public procurement adjudication. The decision not to fine-tune the models for legal analysis was, however, deliberate, reflecting both the substantial temporal and financial costs associated with such domain-specific training.  
Instead, the development effort focused on identifying rulings that addressed the same or closely related substantive issues, following a text-based (linguistic) approach to data extraction rather than a logic-based (symbolic) approach. While this approach facilitated retrieval, it also introduced limitations: semantically similar but contextually unrelated provisions, sometimes drawn from entirely different statutes, could be retrieved if they shared a similar vocabulary. This resulted in hallucinations and improper citations of legal provisions by the models, identified by the committee’s detailed substantive assessment in both parts of the exam. Proper differentiation would require an analysis of the provision’s grouping, the legislative chapter in which it appears, and its intended purpose, none of which was undertaken in this trial. The adopted approach, however, further reflected the differences between the models, resulting from their original characteristics and processing mechanisms, as all of them received the same input while extracting elements from the material in different ways. 

Secondly,  although the team worked closely with one legal practitioner, a more intensive and structured collaboration with multiple lawyers specialising in the relevant legal field could have improved the outcome of the study. Lack of such collaboration was specifically visible in two areas: gaps in domain-specific procedural legal knowledge of the technical team and suboptimal selection of data used to inform the models. The lack of knowledge was visible in the recurring error across all submissions concerning the composition of the adjudicating panel: although the case scenario provided for a single-member panel, all model-generated judgments indicated a three-member panel, suggesting a misapplication of procedural rules. Regarding the composition of the training data, in addition to statutory provisions, the models were also provided with examples of jurisprudence. This decision was based on preliminary trials indicating that the models performed better, according to the technical team, when given access to rulings elaborating on the application of the law, rather than relying solely on statutory text. However, the evaluation committee observed that the knowledge test questions were based primarily on statutory provisions rather than case law. The inclusion of case law may have introduced unnecessary complexity, as synthesizing multiple judicial decisions could have increased the likelihood of errors. This effect was also evident in the practical test, where the models often retrieved and relied on examples of case law that did not necessarily correspond to the specific factual or legal context of the case at hand. Closer interdisciplinary collaboration could improve data selection for model training and reduce knowledge-related errors.

A further limitation concerns evaluator blinding. The team initially planned a fully blinded, ``Turing-style'' assessment, but legal constraints, specifically, that official exam questions could not be disclosed prior to publication of recruitment results, made blinding infeasible. As a result, the assessment committee knew in advance that the analysed outputs were generated by AI. This awareness may have introduced expectancy and confirmation biases \cite{LIM2024100058}, potentially affecting scoring in both the knowledge and practical components. At the same time, the KIO judges maintained that the written works deviated so markedly (and negatively) from their human counterparts that knowing who -- or rather, what -- authored them did not bias their evaluation.

\subsection{Future research}

Building on the limitations above, we see several ways to strengthen validity, robustness, and practical relevance in subsequent studies.

One of the most important areas of research is fine-tuning LLMs to write rulings or judgments (supervised fine-tuning). The training dataset for each training sample should contain the content of the case, fragments of regulations, rulings, and judgments on the basis of which the answer (input) and the content of the judgment (output) should be provided. During inference, fragments of regulations, rulings, and judgments would be searched for automatically. In this context, both the above-mentioned and new, more advanced information-retrieval pipelines would be worth considering. Future work should move beyond purely linguistic similarity toward a structure- and logic-aware stack. Concretely, you could use boundary-aligned, overlapping chunking at legal units. Also, hybrid retrieval methods (sparse + dense) with a law-specific cross-encoder re-ranker are worth exploring. Similarly, the quality of results could be improved by using reinforcement learning, including RLHF. Rewards should be given for correct decisions, correct construction of answers, logical references to regulations and historical judgments, and alignment with the subjective assessment of the expert. It is impossible to construct the training set using only exam cases due to their small number, so it is necessary to consider more advanced methods than the reverse engineering methods presented.

Another interesting direction is to conduct the described experiment (exam) using multi-task agent systems, such as GPT-5 pro, or at least implement them at a specific stage of the process. The design of dedicated, more complex multi-agent systems could also significantly improve results.

Cooperation with legal experts should be deepened. Sustained collaboration between legal experts and technical teams should span task decomposition, dataset curation (including hard negatives), rubric design, and error taxonomy. Embedding domain review into retrieval/ranking development and agent policies can reduce knowledge-related errors (e.g., panel composition) and improve training/evaluation data quality.

Where legally permissible, double-blind setups should be adopted, the presentation and labeling of AI versus human texts should be randomized. These steps will help isolate model performance from expectancy and confirmation biases and improve external validity.

\section*{Ethical Considerations}
All materials used in this study were drawn exclusively from publicly available sources and contained no personally identifiable or otherwise sensitive information. Collaboration with the human legal expert (judge) was conducted in accordance with ethical research practice: participation was voluntary, the nature and purpose of the study were explained in advance, and written informed consent was obtained before any involvement.

The preparation of the document was supported by LLMs, but the design and conduction of the experiments and all findings are genuine.

\section*{Acknowledgments}

***

\bibliographystyle{unsrtnat}
\bibliography{citations}

\newpage
\section*{Appendix}
\begin{appendices}
\renewcommand{\thefigure}{\Alph{figure}}
\footnotesize
\begin{Verbatim}[breaklines=true]
[SYSTEM]
You are an expert legal analyst and senior adjudicator, specializing in Polish law. Your task is to act as an examiner grading a judicial document. You will meticulously evaluate a court verdict document based on the provided rubric. For each criterion, you must assign a score within the specified range and provide a concise, bulleted list of any identified deficiencies or issues that justify your score. Your entire response must be a single, well-formed JSON object.

[CONTEXT]
The user will provide a document containing a court's verdict, a summary of facts, and the judge's reasoning. You are to grade this document with the critical eye of a senior judge reviewing the work of an examinee.

[EVALUATION RUBRIC]
You must evaluate the document against the following criteria and point scales:
1.  **Judgment Construction:** Assesses the formal structure, ensuring all mandatory components of the operative part (the ruling) and the justification (the reasoning) are present and correctly organized. (0-20 points)
2.  **Legal Provisions:** Evaluates the correct selection and interpretation of substantive and procedural law, and its accurate application (subsumption) to the case's established facts. (0-30 points)
3.  **Factual Analysis:** Examines the logical consistency of the factual findings, the soundness of the evidence evaluation, and whether the final decision is a direct and justified consequence of this analysis. (0-30 points)
4.  **Formulation Style (Sentence):** Focuses exclusively on the operative part (the final ruling), assessing its precision, clarity, and legal enforceability. The ruling must be unambiguous. (0-10 points)
5.  **Legal Language:** Assesses the precise and appropriate use of specialized legal terminology and concepts, demonstrating professional legal expertise. (0-5 points)
6.  **Vocabulary:** Measures the fundamental correctness of general and legal vocabulary, including grammar and spelling. (0-2 points)
7.  **Clarity and Conciseness:** Evaluates the ability to express complex reasoning in a clear, concise, and logical manner that is understandable to all parties. (0-2 points)
8.  **Formulation Style (General):** Provides a holistic assessment of the document's overall professionalism, consistency in tone, and stylistic polish. (0-1 point)

[EXAM ANSWERS]

1. **Appeal was raised in time** – The 10-day deadline ends on Saturday, 03.05. However, according to Article 509(2), if the deadline falls on a non-working day, it is extended to the next working day, which is Monday, 05.05.

2. **Proceedings fee too high** – The fee included with the appeal was 20,000 PLN. However, according to §2(1)(2) and §4(4) of the Regulation of the Council of Ministers, the correct fee for this proceeding is 15,000 PLN.

3. **Confirmation of fee being paid was sent in time***

4. **Appeal was correctly signed** - Based on Article 508(1), the document was correctly signed using the trusted profile (profil zaufany).

5. **Interveners were accepted to take part in the trial** - According to Article 525(2), the document regarding intervention in the appeal must be sent before the deadline, not necessarily delivered.

6. **The deadline for the hearing was not kept, but it can be ommitted** - According to Article 544(1), the hearing should have occurred within 15 days of the appeal. However, due to the instructional nature of the deadline (instrukcyjny charakter terminu), this does not result in procedural consequences.

7. **Number of committee members is correct** - According to Article 487(2), the President of the Chamber (Prezes Izby) may order the number of committee members to be reduced to one.

8. **The appeal is to be settled in favor of appellant** - The defendant is to revoke their selection of the best offer and the rejection of the appellant’s offer, re-do the process of offer analysis and selection, and request the appellant to submit the missing documents based on Article 107(2).

9. **Distribution of costs** - The cost of the proceedings is 15,000 PLN, the cost of legal representation (pełnomocnik) is 3,600 PLN, and the cost of travel is 160 PLN, totaling 18,760 PLN. These costs should be paid by the defendant to the benefit of the appellant, based on §5(2)(a) of the Regulation of the Council of Ministers dated 30 December 2020.

10. **Defendant has the right to defend** - The fact that the defendant (buyer) did not respond to the appeal does not revoke their right to be represented at the hearing.

11. **Appellant's incorrect document is not considered to be the fault of appellant** - In the evidence provided, the appellant’s document was incomplete. However, since the defendant (buyer) declared that their specifications could be updated, which led to the deficiency in the appellant’s document, based on Article 107(2), the defendant should have requested the appellant to update their document.

12. **Faulty document was not relevant to defendant's criteria** - The appellant’s faulty document did not contain matters directly relevant to the defendant’s criteria for evaluating the offer. The document containing those criteria was not faulty, therefore the defendant should not have rejected the offer based on the faulty document.

[INSTRUCTIONS]
1.  Thoroughly and critically read the entire provided legal document.
2.  Perform an evaluation of the document based on the information in `[EXAM ANSWERS]`. Assess whether the reasoning, conclusion and the required legal provisions were included in the document.
3.  For each criterion in the `[EVALUATION RUBRIC]`, conduct a detailed assessment of the document's content and style.
4.  Assign a point value for each criterion, ensuring it is within the specified range.
5.  For each criterion, compile a list of specific, actionable issues or deficiencies found in the text. This list serves as the justification for your score. If no issues are found for a criterion, provide an empty list `[]`.
6.  Construct the final JSON output according to the `[OUTPUT FORMAT]`. Do not include any text or explanation outside of the final JSON object.

[OUTPUT FORMAT]
Your entire output must be a single JSON object with the following structure:
{
  "judgment_construction": {
    "points": <integer>,
    "issues": [
      "<string issue 1>",
      "<string issue 2>"
    ]
  },
  "legal_provisions": {
    "points": <integer>,
    "issues": []
  },
  "factual_analysis": {
    "points": <integer>,
    "issues": [
      "<string issue 1>"
    ]
  },
  "formulation_style_overall": {
    "points": <integer>,
    "issues": []
  },
  "legal_language": {
    "points": <integer>,
    "issues": [
      "<string issue 1>"
    ]
  },
  "vocabulary": {
    "points": <integer>,
    "issues": []
  },
  "clarity_and_conciseness": {
    "points": <integer>,
    "issues": [
      "<string issue 1>",
      "<string issue 2>"
    ]
  },
  "formulation_style_specific": {
    "points": <integer>,
    "issues": []
  }
}

[DOCUMENT TO BE EVALUATED]
\end{Verbatim}
\par\addvspace{\medskipamount}
\refstepcounter{figure}
\small
\noindent\textbf{Fig. \thefigure} LLM Prompt for written exam assessment -- original English version. The English version was used for all LLMs.
\footnotesize
\label{listing:judge-test-prompt}
\par\addvspace{\medskipamount}

\begin{Verbatim}[breaklines=true]
{
  "judgment_construction": {
    "points": 15,
    "issues": [
      "The ruling and justification sections are present, but the structure is somewhat disorganized with sections blending into one another without clear distinctions.",
      "The numbering and titling of sections in the document are inconsistent, causing some confusion in navigation for the reader."
    ]
  },
  "legal_provisions": {
    "points": 28,
    "issues": [
      "There is a minor discrepancy in referencing Article 107. The document mentions 107 ust. 3 in the context of not supplementing documents, though Article 107 ust. 2 should have been evaluated concerning the appellant's claim."
    ]
  },
  "factual_analysis": {
    "points": 25,
    "issues": [
      "While the factual findings are generally consistent, there is insufficient explanation of how specific evidence was weighed, particularly regarding the appellant's claims about the document sufficiency."
    ]
  },
  "formulation_style_overall": {
    "points": 1,
    "issues": []
  },
  "legal_language": {
    "points": 4,
    "issues": [
      "Minor technical inaccuracies in the usage of legal articles and language consistency could be improved, including the switching of 'SWZ' terms without prior clear explanation."
    ]
  },
  "vocabulary": {
    "points": 2,
    "issues": []
  },
  "clarity_and_conciseness": {
    "points": 1,
    "issues": [
      "The document could benefit from more concise language; some sections are verbose, affecting the clarity of the analysis.",
      "Certain repetitive statements about 'głębokości brodzenia' add unnecessary bulk."
    ]
  },
  "formulation_style_specific": {
    "points": 9,
    "issues": []
  }
}
\end{Verbatim}
\par\addvspace{\medskipamount}
\refstepcounter{figure}
\small
\noindent\textbf{Fig. \thefigure} Result of the evaluation of Bielik-11B-v2.6 by GPT-4o.
\footnotesize
\label{listing:bielik-llm-eval}
\par\addvspace{\medskipamount}

\begin{Verbatim}[breaklines=true]
{
  "judgment_construction": {
    "points": 18,
    "issues": [
      "The document contains most mandatory components, but the organization of some sections could be clearer and more logically structured."
    ]
  },
  "legal_provisions": {
    "points": 28,
    "issues": [
      "Some legal provisions are not fully elaborated, though correctly referenced."
    ]
  },
  "factual_analysis": {
    "points": 27,
    "issues": [
      "The analysis could be more detailed in linking facts directly to legal conclusions."
    ]
  },
  "formulation_style_overall": {
    "points": 9,
    "issues": []
  },
  "legal_language": {
    "points": 4,
    "issues": [
      "While specialized terminology is used, the integration of some terms could be smoother for better readability."
    ]
  },
  "vocabulary": {
    "points": 2,
    "issues": []
  },
  "clarity_and_conciseness": {
    "points": 1,
    "issues": [
      "Some sections are overly detailed, leading to reduced clarity.",
      "A more concise summary of the legal reasoning could improve understandability."
    ]
  },
  "formulation_style_specific": {
    "points": 1,
    "issues": []
  }
}
\end{Verbatim}

\par\addvspace{\medskipamount}
\refstepcounter{figure}
\small
\noindent\textbf{Fig. \thefigure} Result of the evaluation of Claude 4 Sonnet by GPT-4o.
\footnotesize
\label{listing:claude-llm-eval}
\par\addvspace{\medskipamount}

\begin{Verbatim}[breaklines=true]
{
  "judgment_construction": {
    "points": 15,
    "issues": [
      "Missing clear separation and labeling of sections within the document.",
      "Title of each section like 'Factual Findings', 'Legal Analysis', etc., should be more explicitly demarcated."
    ]
  },
  "legal_provisions": {
    "points": 28,
    "issues": [
      "The justification for the decision lacks explicit references to how specific legal principles were applied."
    ]
  },
  "factual_analysis": {
    "points": 27,
    "issues": [
      "Some assumptions about the necessity of specific document pages were not thoroughly explained in the context of the objective criteria."
    ]
  },
  "formulation_style_overall": {
    "points": 1,
    "issues": []
  },
  "legal_language": {
    "points": 4,
    "issues": [
      "Some legal terms are assumed to be known without definitions or context for clarity to all readers."
    ]
  },
  "vocabulary": {
    "points": 2,
    "issues": []
  },
  "clarity_and_conciseness": {
    "points": 1,
    "issues": [
      "Long and complex sentences reduce the clarity of the reasoning.",
      "Some paragraphs are densely packed with information without adequate summarization."
    ]
  },
  "formulation_style_specific": {
    "points": 10,
    "issues": []
  }
}
\end{Verbatim}

\par\addvspace{\medskipamount}
\refstepcounter{figure}
\small
\noindent\textbf{Fig. \thefigure} Result of the evaluation of GPT 4.1 by GPT-4o
\footnotesize
\label{listing:gpt-llm-eval}
\par\addvspace{\medskipamount}

\begin{figure}[h]
\footnotesize
\textbf{appealDate -- Original prompt}
\begin{Verbatim}[breaklines=true]
System: IMPORTANT: For your response to user, you MUST use the `format_final_json_response` tool with your complete answer formatted according to the required schema. Do not attempt to format the JSON manually - always use this tool. Your response will be rejected if it is not properly formatted through this tool. Only use this tool once you are ready to provide your final answer.
Human: # Wyodrębnij datę, w której Odwołujący wniósł odwołanie:

```
{Description of the facts in Polish}
```
\end{Verbatim}

\textbf{appealDate -- Translated prompt}
\begin{Verbatim}[breaklines=true]
System: IMPORTANT: For your response to user, you MUST use the `format_final_json_response` tool with your complete answer formatted according to the required schema. Do not attempt to format the JSON manually - always use this tool. Your response will be rejected if it is not properly formatted through this tool. Only use this tool once you are ready to provide your final answer.
Human: # Select the date on which the Appellant filed the appeal:

```
{Description of the facts in Polish}
```
\end{Verbatim}

\caption{Prompt used to extract appealDate from the factual description. The part that can be called the prompt system is the default content of the n8n framework. It can be modified, but in this case we did not consider it necessary. A JSON Schema was provided with the model along with the prompt. The description of the facts provided at the input was put in the "Description of the facts in Polish" placeholder.}
\label{listing:appeal-date-prompt}
\end{figure}

\begin{figure}[t]
\footnotesize
\textbf{appealDate -- Original JSON Schema}
\begin{Verbatim}
{
  "type": "object",
  "properties": {
    "appealDate": {
      "type": "string",
      "description": "Data, w której Odwołujący wniósł odwołanie. Format 
      daty to YYYY-MM-DD, inaczej rok-miesiąc-dzień, czyli np 2023-12-27, 
      a nie 27-12-2023. Jeśli data nie pasuje do formatu YYYY-MM-DD, 
      przekonwertuj ją na ten format."
    },
    "explanation": {
      "type": "string",
      "description": "Uzasadnienie decyzji"
    }
  },
  "required": [
    "appealDate"
  ]
}
\end{Verbatim}

\textbf{appealDate -- Translated JSON Schema}
\footnotesize
\begin{Verbatim}
{
  "type": "object",
  "properties": {
    "appealDate": {
      "type": "string",
      "description": "The date on which the appellant lodged the appeal. 
      The date format is YYYY-MM-DD, i.e. year-month-day, e.g. 2023-12-27, 
      not 27-12-2023. If the date does not match the YYYY-MM-DD format, 
      convert it to this format."
    },
    "explanation": {
      "type": "string",
      "description": "Reasons for the decision"
    }
  },
  "required": [
    "appealDate"
  ]
}
\end{Verbatim}

\caption{JSON Schema used to extract appealDate from text.}
\label{listing:appeal-date-json-schema}
\end{figure}

\begin{figure}[t]
\footnotesize
\textbf{expectedFee -- Original prompt}
\begin{Verbatim}[breaklines=true]
System: IMPORTANT: For your response to user, you MUST use the `format_final_json_response` tool with your complete answer formatted according to the required schema. Do not attempt to format the JSON manually - always use this tool. Your response will be rejected if it is not properly formatted through this tool. Only use this tool once you are ready to provide your final answer.
Human: # Zadanie
Na podstawie fragmentu rozporządzenia zadecyduj oraz stanu faktycznego zdecyduj, ile powinien wynosić wpis (opłata) od odwołania. Nie uwzględniaj informacji, ile rzeczywiście wniósł. Badam, czy wniesiona opłata była wystarczająca.

# Ważne informacje
a) Zamówienie jest o wartości powyżej lub równej progom unijnym.
b) opracowanie dokumentacji projektowej, projektu, planu itp ale bez budowy, to usługa, a nie roboty budowlane
c) zwrot "zaprojektuj i buduj", "zaprojektuj i wybuduj" zawsze oznacza roboty budowlane
d) jeśli budowa jest tylko niewielkim elementem zamówienia, a składa się na niego więcej różnych usług i dostaw, to rostrzygaj to jako usługi lub dostawy.

# Uproszczone rozporządzenie
```markdown
{A simplified version of § 2 of the Regulation of the Minister of Finance}
```

# Stan faktyczny
```markdown
{Description of the facts in Polish}
```
\end{Verbatim}

\caption{Prompt used to extract expectedFee from the factual description. As in Figure \ref{listing:appeal-date-prompt}, The part that can be called the prompt system is the default content of the n8n framework. The description of the facts provided at the input was put in the "Description of the facts in Polish" placeholder. A simplified version of § 2 of the Regulation of the Minister of Finance can be seen in both Figure \ref{listing:simplified-regulation-original} and \ref{listing:simplified-regulation}}
\label{listing:expected-fee-prompt-original}
\end{figure}

\begin{figure}[t]
\footnotesize
\textbf{expectedFee -- Translated prompt}
\begin{Verbatim}[breaklines=true]
System: IMPORTANT: For your response to user, you MUST use the `format_final_json_response` tool with your complete answer formatted according to the required schema. Do not attempt to format the JSON manually - always use this tool. Your response will be rejected if it is not properly formatted through this tool. Only use this tool once you are ready to provide your final answer.
Human: # Task
Based on the fragment of the regulation and the facts, decide how much the entry (fee) for the appeal should be. Do not take into account how much he actually paid. I am examining whether the fee paid was sufficient.

# Important information
a) The contract value is above or equal to the EU thresholds.
b) The development of project documentation, design, plan, etc., but without construction, is a service and not construction work.
c) The phrases “design and build” and “design and construct” always refer to construction works.
d) If construction is only a minor part of the contract and it consists of more diverse services and supplies, treat it as services or supplies.

# Simplified regulation
```markdown
{A simplified version of § 2 of the Regulation of the Minister of Finance}
```

# Facts
```markdown
{Description of the facts in Polish}
```
\end{Verbatim}

\caption{Translated version of the expectedFee prompt}
\label{listing:expected-fee-prompt-translated}
\end{figure}

\begin{figure}[t]
\footnotesize
\textbf{expectedFee -- A simplified version of § 2 of the Regulation of the Minister of Finance (original)}
\begin{Verbatim}[breaklines=true]
## 1. Wpis w postępowaniu na dostawy, usługi lub konkurs: wpis **15000 zł**

---

## 2. Wpis w postępowaniu na roboty budowlane: wpis **20 000 zł**

---

## 3. Wpis w postępowaniu na usługi społeczne i inne usługi szczególne: wpis **15000 zł**

---

## 4. Kiedy sprawa toczy się o to, że Zamawiający powinien był zorganizować postępowanie lub konkurs, a go nie zorganizował.

* **Dostawy, usługi lub konkurs** → wpis **7500 zł**
* **Roboty budowlane** → wpis **10000 zł**

---

## 5. Jednolity wpis przy ofertach częściowych

Jeśli odwołujesz się do czynności lub zaniechania dotyczącego kilku części zamówienia, wnosisz **jeden wpis**.
\end{Verbatim}

\caption{The original version of the simplified regulation. The translated version of the prompt is available in Figure~\ref{listing:simplified-regulation}.}
\label{listing:simplified-regulation-original}
\end{figure}

\renewcommand{\thefigure}{\arabic{figure}}

\end{appendices}

\end{document}